\PassOptionsToPackage{table}{xcolor}
\documentclass[10pt, a4paper, logo]{qwen}
\usepackage{wrapfig} 
\usepackage{times}
\usepackage{amsmath}
\usepackage{amssymb}
\usepackage{booktabs}
\usepackage{graphicx}
\usepackage{svg}
\usepackage{subcaption}
\usepackage{float}
\usepackage{multirow}
\usepackage{array}
\usepackage{tabularx}
\usepackage{xspace}
\usepackage[table]{xcolor}
\usepackage{siunitx}
\usepackage[ruled,linesnumbered,lined]{algorithm2e}
\usepackage{enumitem}
\usepackage{natbib}
\usepackage[most]{tcolorbox}
\newtheorem{lemma}{Lemma}
\newtheorem{corollary}{Corollary}
\newtheorem{theorem}{Theorem}

\newcolumntype{C}[1]{>{\centering\arraybackslash}p{#1}}
\newcolumntype{L}[1]{>{\raggedright\arraybackslash}p{#1}}
\newcolumntype{Y}{>{\centering\arraybackslash}X}
\definecolor{TailSieveLight}{HTML}{EAF3FF}
\definecolor{TailSieveDeep}{HTML}{BDD7F7}
\newcommand{\method}{\textsc{TailSieve}\xspace}

\newtcolorbox{promptbox}[1]{%
    enhanced,
    breakable,
    colback=gray!7,
    colframe=gray!35,
    boxrule=0.45pt,
    arc=1.5pt,
    left=7pt,
    right=7pt,
    top=5pt,
    bottom=5pt,
    before skip=5pt,
    after skip=8pt,
    title=#1,
    colbacktitle=gray!16,
    coltitle=black,
    fonttitle=\small\bfseries\sffamily,
    fontupper=\small\ttfamily,
}

\newcommand{\teaserfigure}{%
    \begin{center}
        \includegraphics[width=0.90\textwidth]{figures/main.pdf}
        \captionof{figure}{Overview of \method (left) and
        routing-with-speculation speedups on two Qwen3.5 models (right).}
        \label{fig:teaser}
    \end{center}
}
\let\qwenabscontent\abscontent
\renewcommand{\abscontent}{\teaserfigure\qwenabscontent}

\title{\textsc{TailSieve}: Partial-Rollout-Guided Tail Routing for LLM Rollouts}

\author[1,2]{Tianqi Xu$^{*}$}
\author[1]{Lu Lv$^{*}$}
\author[1,3]{Haoyang Huang$^{*}$}
\author[1,3]{Wenjie Huang$^{*}$}
\author[3]{Zhanming Shen}
\author[3]{Yuhao Shen}
\author[1]{Baolin Zhang}
\author[1]{Xinyi Hu\textsuperscript{\dag}}
\author[1]{Shuang Ge}
\author[1]{Jun Dai}
\author[1]{Tianyu Liu}
\author[4]{Suorong Yang}
\author[5]{Zhikai Li}
\author[1]{Ye Bai}
\author[1]{Jun Zhang\textsuperscript{\dag}}
\author[1]{Lei Chen}
\author[1]{Yue Li}
\author[1]{Mingchen Wan}
\affil[1]{Qwen Business Unit of Alibaba}
\affil[2]{Carnegie Mellon University}
\affil[3]{Zhejiang University}
\affil[4]{National University of Singapore}
\affil[5]{Institute of Automation, Chinese Academy of Sciences}

\begin{abstract}
Large-scale rollouts have become a core component of modern LLM systems, spanning reinforcement learning (RL) post-training, on-policy distillation, and sampling-heavy evaluation. Unlike online serving, which typically targets request-level latency and throughput, synchronous rollouts must collect a complete batch before proceeding. A few long-tail generations can therefore dominate an entire rollout step, and uniform routing can slow these generations further by placing them in high-concurrency decoding batches.

We present \method, a partial-rollout-guided framework that jointly controls tail routing and replica allocation. Motivated by the observed persistence of prompt-level tail tendencies across policy updates, \method uses partial-rollout completion status as a training-free signal for identifying candidate tail groups. Selected prompts are combined with newly sampled prompts, and all responses are generated from scratch under the current policy. A hierarchical controller uses response-work history and a serving model: a fast inner loop adjusts the number of isolated groups to balance completion times, while a slower outer loop adapts how many replicas serve these groups. The resulting low-concurrency execution also enables route-specialized speculative decoding with MTP or DFlash. Across the evaluated configurations, \method achieves up to 1.67$\times$ speedup with routing alone and 2.59$\times$ with routing and speculation combined, both over uniform group routing.
\end{abstract}

\begin{document}
\maketitle

\section{Introduction}

Large-scale rollouts support reinforcement learning with verifiable
rewards~\citep{guo2025deepseekr1}, on-policy distillation
\citep{agarwal2024opd}, synthetic data generation, and sampling-intensive
evaluation~\citep{wang2023selfinstruct,chen2021codex}. These pipelines
generate hundreds or thousands of responses per step, making rollout
generation a dominant component of RL training time
\citep{zhouapril2025}.

Unlike online serving systems, where the primary objectives are
request-level latency, throughput, and cost efficiency
~\citep{qiu2024ssjf,liu2026dualpool,yuan2026dualmap,
hu2026echo}, synchronous rollout pipelines are governed by a different
performance objective: step-level makespan. The optimization process
cannot proceed until a required set of generations is completed.
Therefore, the completion time of a rollout step is determined by the
slowest unfinished requests (Figure~\ref{fig:decode-throughput}(b)). Since LLM response lengths exhibit strong
skewness, a small number of extremely long generations can dominate the
overall execution time and stall the entire rollout
process~\citep{zhouapril2025,zhong2025streamrl}.

Existing works improve rollout efficiency by reducing synchronization stalls,
reorganizing long-tail workloads, adapting resource allocation, or accelerating
decoding~\citep{zhouapril2025,gao2026rollpacker,zhong2025streamrl,
qin2026seer,shao2026das,xububblespec2026,chen2026libra}.
Appendix~\ref{app:related-work} compares these approaches.
Here, we focus on a fundamental routing problem in synchronous rollouts:
\textbf{How should requests be distributed across replicas to minimize step-level makespan?}

To answer this question, we begin by studying the offline optimal assignment, assuming that all response lengths are known in advance.
In the evaluated long-tail cases, the optimum follows a \emph{tail-isolation-with-balancing}
structure: it assigns a small number of extreme-tail requests to one replica, which
we call the tail replica, while routing most requests to the other, or bulk replica (Figure~\ref{fig:optimal-routing}(a)). We further find a simpler and more tractable solution: isolating the \(k\) longest responses on the tail replica and routing the rest to the bulk replica can closely match the optimal makespan (Figure~\ref{fig:optimal-routing}(b); Appendix~\ref{app:offline-routing-simulation}).

Applying this top-\(k\) strategy online to reduce rollout makespan requires identifying likely tail prompts in advance. The gradual evolution of the policy during RL training suggests a possible way forward: a prompt's behavior in one rollout may help identify whether it will generate long responses in the next. Motivated by this possibility, we examine response lengths across checkpoints. We find that prompt identity accounts for most of the variation (Appendix~\ref{app:length-variance}): prompts that generate long responses tend to remain near the tail after policy updates (Figure~\ref{fig:prompt-tail-auc}). We also find that groups unfinished at a partial-rollout cutoff capture most of the longest-response groups (Figure~\ref{fig:partial-filter}).

\begin{figure}[t]
    \centering
    \includegraphics[width=\linewidth]{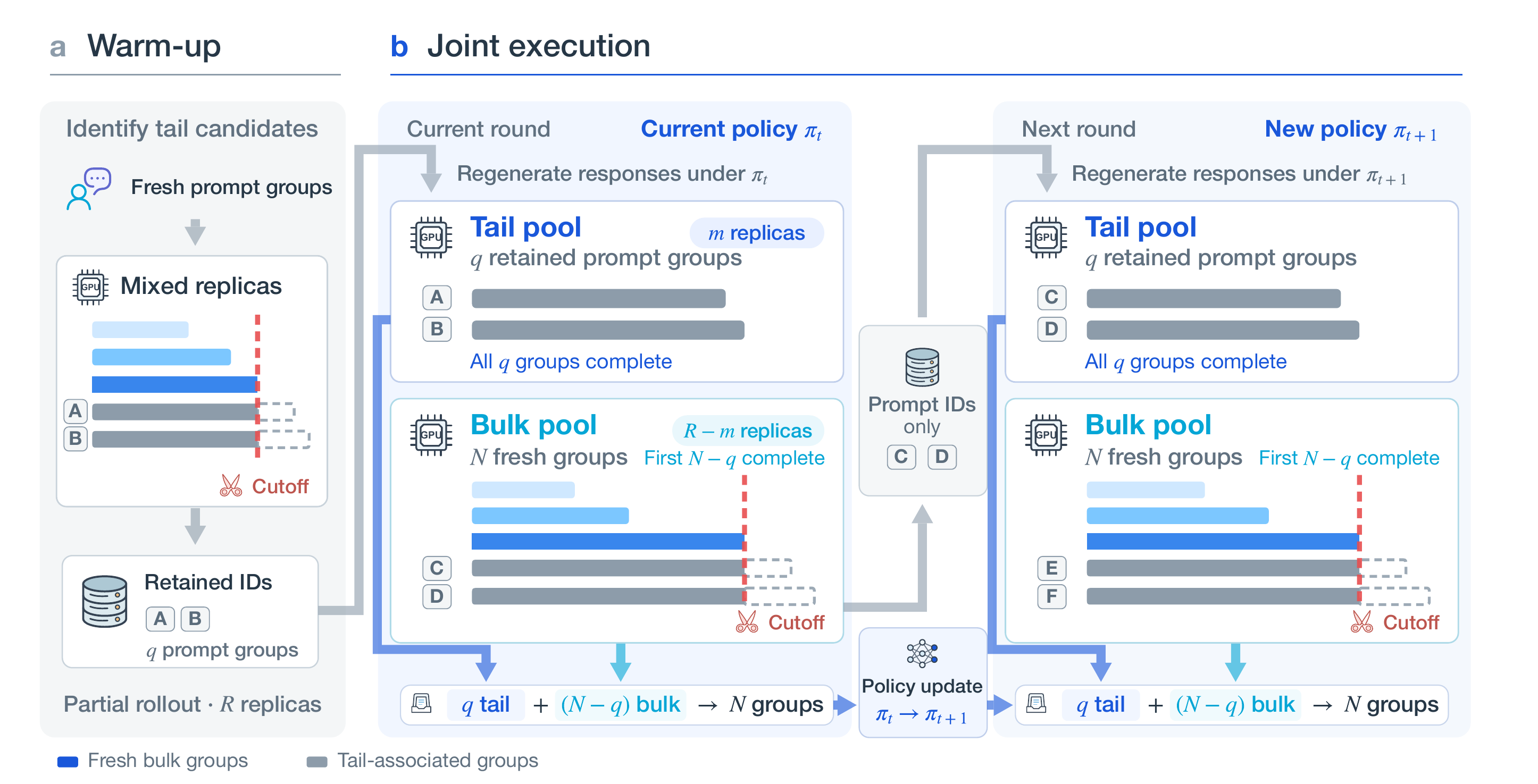}
    \caption{\method warm-up and joint execution. Unfinished prompt
    identities enter the tail pool and regenerate under the current
    policy; generated prefixes and KV caches are not carried across
    policy updates.}
    \label{fig:method-overview}
\end{figure}

These observations motivate \method's use of partial-rollout status as a prompt-level tail signal. It routes unfinished prompt groups---likely to be the top-\(k\) longest groups---to dedicated tail replicas in the next round, regenerating their responses under the current policy (Figure~\ref{fig:method-overview}).
\method requires no auxiliary length model and does not carry unfinished trajectories across policy updates. An online controller jointly adapts tail workload and replica capacity (Section~\ref{sec:allocation-config}), while each update mixes regenerated tail prompts with newly sampled prompts. Empirically, accepted-group mean-length distributions remain close to those under uniform routing (Figure~\ref{fig:deepscaler-group-length-cdf}).

Figure~\ref{fig:teaser} summarizes the design and headline results.
Our contributions are summarized as follows:

\begin{itemize}[itemsep=0pt, topsep=0pt, leftmargin=19pt]
\item[(1)] \textit{Long-Tail-Aware Routing Principles}: We formulate long-tail-aware rollout routing as a step-level
makespan minimization problem. In the long-tail regime, the offline
optimum follows \textbf{tail isolation with load balancing}, and a simple
top-\(k\) policy closely approximates the exact oracle
(Section~\ref{sec:optimal-routing}; Figure~\ref{fig:optimal-routing}).
\item[(2)] \textit{Adaptive Hierarchical Routing}: We develop \method, which uses partial rollout as a training-free
prompt-level tail signal. Its hierarchical controller jointly adapts the
isolated workload and replica split using pool completion times,
response-work history, and measured decoding throughput
(Section~\ref{sec:allocation-config}; Algorithm~\ref{alg:tailsieve-controller}).
\item[(3)] \textit{Tail Isolation Amplifies Speculative Decoding}: We show
that tail isolation amplifies the gains from speculative decoding by
concentrating long responses on low-concurrency tail replicas.
This structure enables \textbf{deeper speculation} and extends its
beneficial operating range, which \method exploits through
route-specialized proposal depths to further reduce rollout makespan
(Figure~\ref{fig:spec-convergence}(a)). Across five models and two
workloads, routing alone reaches 1.67$\times$ speedup and routing with
speculation reaches 2.59$\times$, both over uniform group routing.
\end{itemize}

\section{Motivating Observations}
\label{sec:optimal-routing}
\label{sec:tail-isolation}
We begin with a simple question: \textbf{if we knew which generations were going to be long, how should they be routed across replicas?}

\begin{figure}[t]
\centering
\includegraphics[width=\linewidth,trim=0 0 0 140bp,clip]{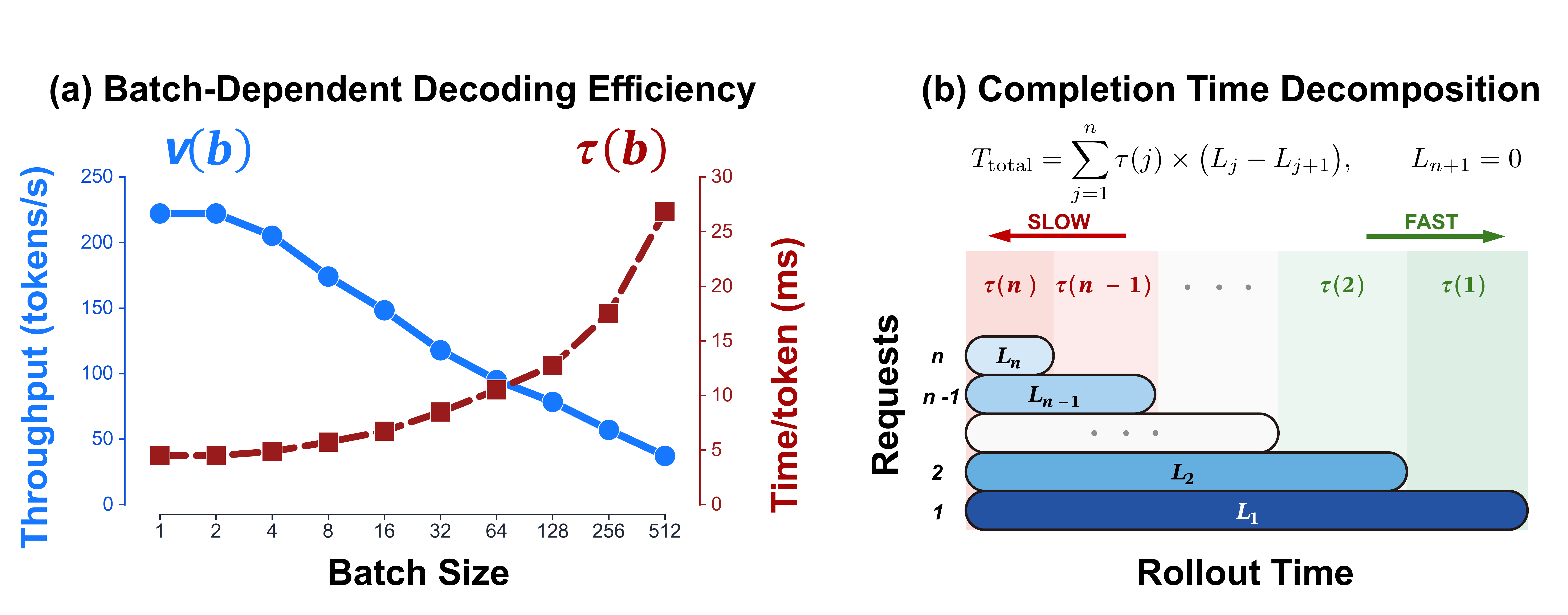}
\caption{Measured per-request decode throughput versus active batch size (a) and the resulting replica-time decomposition (b). The left axis reports tokens/s per request, and the right axis reports ms/token. The objective follows the changing concurrency until the last request finishes.}
\label{fig:decode-throughput}
\end{figure}

\paragraph{A decoding-time objective.}
We first consider a simple model that focuses only on decoding. Let $v(b)$ be the per-request throughput at active batch size $b$, and let $\tau(b)=1/v(b)$. For a replica with request lengths $L_1 \ge \cdots \ge L_n$ and $L_{n+1}=0$, its completion time is

\begin{equation}
T(S)
=
\sum_{j=1}^{n}\tau(j)(L_j-L_{j+1})
=
\sum_{j=1}^{n}L_j[\tau(j)-\tau(j-1)],
\qquad \tau(0)=0.
\label{eq:decoding-time}
\end{equation}

The routing objective is to minimize the maximum completion time across replicas. Figure~\ref{fig:decode-throughput} illustrates the decoding model.

\paragraph{Exact oracle and tail isolation.}
We process requests from longest to shortest. Assigning a request of length $L_i$ to a replica with $c$ existing requests increases its completion time by $L_i[\tau(c+1)-\tau(c)]$. A dynamic program tracks the completion times of the two replicas and prunes dominated states.
Appendix~\ref{app:oracle-proof} specifies the recurrence and its exactness
under the decoding model. In these 100-request instances,
Pareto pruning~\citep{ehrgott2005multicriteria} reduces the naive
$2^{100}$ assignments to roughly $10^6$--$10^7$ retained states.
This computation provides an offline reference for evaluating online routing.

\suppressfloats[t]
\begin{figure}[t]
\centering
\includegraphics[width=0.95\linewidth]{figures/optimal_routing.pdf}
\caption{A representative offline assignment and top-$k$ isolation sweep from the original rollout analysis. The oracle uses few tail requests plus balancing traffic; top-$k$ is a simpler restricted policy.}
\label{fig:optimal-routing}
\end{figure}

In the representative 100-request case (Figure~\ref{fig:optimal-routing}(a)), the optimal assignment is highly imbalanced: the tail lane contains only four responses---the two longest responses together with two additional responses for load balancing---yet takes nearly the same time to finish as the bulk lane with the remaining 96 responses. A simpler restricted policy isolates the $k$ longest responses and closely matches the optimal makespan (Figure~\ref{fig:optimal-routing}(b)). Across 100 sampled cases, the best-$k$ solution is within 4\% of the optimum in all cases and within 1\% in 72\% of cases (Appendix~\ref{app:offline-routing-simulation}).

The representative best-$k$ assignment takes 54.84\,s, recovering most
of the oracle improvement without solving the full assignment problem.

\subsection{Partial Rollout as a Cross-Round Tail Signal}
\label{sec:partial-rollout}
Although completion lengths are unknown online, the top-$k$ isolation result provides a useful insight: we only need to identify the longest responses in each batch, rather than predict their exact lengths.

\begin{figure}[!b]
\centering
\includegraphics[width=0.95\linewidth]{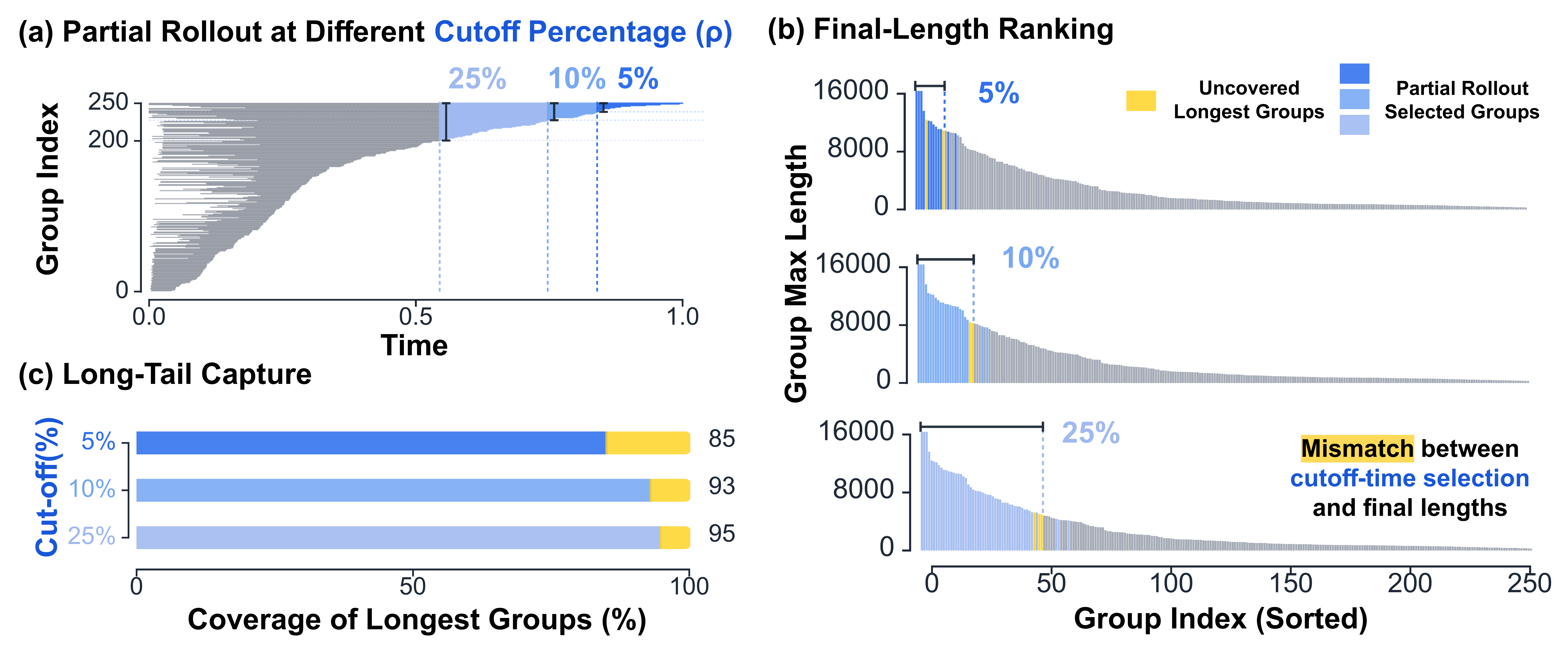}
\caption{Partial-rollout tail identification at prompt-group granularity.
The cutoff parameter $\rho$ is the number of excess groups divided by the
completion quota. Appendix~\ref{app:partial-cutoff} defines the virtual
cutoff, group-length ranking, and coverage calculation.}
\label{fig:partial-filter}
\end{figure}

Unlike prior systems that carry or reuse unfinished trajectories across rollout steps~\citep{zhouapril2025,kimi2025k15}, \method uses cutoff-time status only as a prompt-level routing signal. Figure~\ref{fig:partial-filter} evaluates this signal at prompt-group granularity, comparing groups unfinished at cutoffs with the final group-length ranking. We further conduct a cross-checkpoint study and find that, even as the model evolves during training, prompt-level length rankings remain highly informative for tail identification: adjacent-checkpoint AUC ranges from 0.957 to 0.985. Our cross-checkpoint study finds that prompt identity accounts for most
of the response-length variance (Appendix~\ref{app:prompt-stability}).
At excess-group ratios $\rho=5\%,10\%,25\%$, the filter captures
$85\%,93\%,95\%$ of the corresponding length-ranked tail groups;
coverage approaches $100\%$ for more extreme tail subsets.
Appendix~\ref{app:partial-cutoff} defines the retrospective cutoff and
coverage calculation.

\subsection{Coupled Tail-Workload and Replica Allocation}
\label{sec:conditional-gain}
\begin{figure}[htbp]
    \centering
    \includegraphics[width=0.95\linewidth]{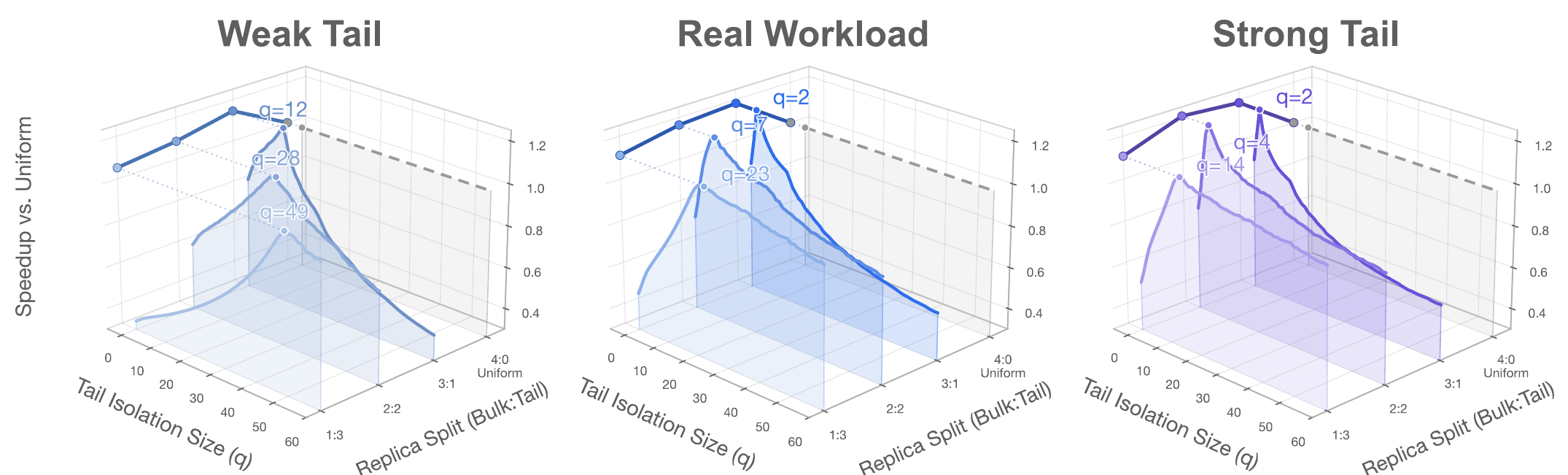}
    \caption{Joint-control landscapes showing that the optimal isolation
    size and replica split vary with tail strength. Strengthening the
    response-length tail moves the preferred operating point toward a smaller
    isolation size $q$, whereas weakening the tail requires a larger $q$ to
    keep the two pools balanced and brings the optimum closer to uniform
    routing.}
    \label{fig:conditional-gain}
\end{figure}

Our remaining challenge is deciding how much tail workload should be isolated. Intuitively, when a small number of responses are substantially longer than the rest, only these extreme responses should be routed separately; when response lengths are nearly uniform, tail isolation provides little benefit. As the response-length distribution shifts during training, the optimal operating point therefore shifts as well. Figure~\ref{fig:conditional-gain} illustrates this relationship between workload shape and the preferred tail configuration. We provide a detailed analysis in Appendix~\ref{app:conditional-gain} showing that the optimal operating point depends jointly on the response-length distribution and the concurrency-dependent decoding throughput.

This workload-dependent operating point motivates \method: rather than using a fixed tail fraction or replica split, \method adapts both the amount of isolated tail workload and the resources assigned to it according to the current response-length distribution. The request-level oracle further reveals that the key structural benefit comes from isolating long-tail work. \method translates this oracle insight into an online, deployable mechanism by identifying tail-prone prompt groups and routing them to dedicated replicas.

\section{\method Design}
\label{sec:method}

\paragraph{A mixed tail--bulk pipeline.}
\method converts the routing structure of Section~\ref{sec:optimal-routing}
into a cross-round pipeline. A partial rollout identifies unfinished prompt groups as tail candidates. In the next step, the tail pool regenerates these prompts under the current policy, while the bulk pool processes fresh groups. Only prompt identities are carried across steps.

At a stable allocation, the tail pool uses $m_t$ of $R$ replicas to regenerate $q_t$ carried groups; bulk launches $N$ fresh groups on the rest. The update accepts all $q_t$ tail groups and the first $N-q_t$ completed bulk groups, totaling $N$. Unaccepted bulk groups form the next carry. Thus $q_t$ controls isolated workload and $m_t$ tail capacity; $(q_t,m_t)=(0,0)$ gives uniform routing.

Fresh regeneration ensures that all responses are sampled under the current
policy. Under the idealized assumptions in
Appendix~\ref{app:length-distribution}, the mixed tail--bulk batch preserves
the response-length distribution in expectation for a fixed policy.
Empirically, the accepted-group mean-length distributions remain close to
those under uniform routing
(Appendix~\ref{app:accepted-group-length-distributions}).
Overall, \method exploits persistent prompt-level tail signals for efficient
routing while keeping every response on-policy.

\suppressfloats[t]

\subsection{Hierarchical Allocation}
\label{sec:allocation-config}

For each rollout step, execution completes only when both the tail and bulk pools finish, so the slower pool determines the step time:
\begin{equation}
T_t(q,m)=\max\!\left\{
T_t^{\mathrm{tail}}(q,m),
T_t^{\mathrm{bulk}}(N-q,R-m)
\right\}.
\label{eq:controller-objective}
\end{equation}
Here, $T_t^{\mathrm{bulk}}(N-q,R-m)$ denotes the time to collect $N-q$ complete groups from the $N$ groups submitted to bulk.
A fast inner loop adjusts the tail quota $q$ from measured completion times,
while a slower outer loop uses a profiled throughput curve and the
empirical completed-token work profile to update the replica split $m$.

\paragraph{Inner loop: balance completion times.}
For a fixed replica split, the min--max objective is minimized when both pools
finish together (Appendix~\ref{app:inner-balance}). Eligible rounds give the
group-processing rates
\begin{equation}
T^{\mathrm{tail}} \approx \frac{q}{r^{\mathrm{tail}}},
\qquad
T^{\mathrm{bulk}} \approx \frac{N-q}{r^{\mathrm{bulk}}}.
\end{equation}
Equating the two gives the balanced quota
\begin{equation}
q_{\mathrm{balanced}}
=
N\frac{\bar r^{\mathrm{tail}}}
{\bar r^{\mathrm{tail}}+\bar r^{\mathrm{bulk}}},
\label{eq:inner-control}
\end{equation}
where $\bar r$ is an exponential moving average. The controller projects and
clips this target, then selects the better of its floor and ceiling under the
same min--max surrogate. The core rule stops when the mean gap is within
$\epsilon_q$; Appendix~\ref{app:controller-calibration} gives the measurement
validity checks, finite-integer safeguards, and parameter settings used in the
implementation.

\paragraph{Outer loop: adjust replica capacity.}
After balancing $q$, a throughput curve and the empirical work profile
score the current and feasible adjacent splits after rebalancing their quota:
\begin{equation}
\widehat\Phi(m')=\min_{q'\in\mathcal Q}
\max\!\left\{\widehat T^{\mathrm{tail}}(q',m'),
\widehat T^{\mathrm{bulk}}(q',m')\right\}.
\label{eq:outer-model-objective}
\end{equation}
The controller moves one replica only when the best neighbor improves this
score by more than $\epsilon_m$, without timing candidate layouts online.
Appendix~\ref{app:controller-calibration} gives the implementation;
Appendix~\ref{app:outer-local-search} gives the fluid-limit support for
neighboring-split search.

\begin{algorithm}[t]
\footnotesize
\DontPrintSemicolon
\KwIn{Groups $N$, replicas $R$, feasible quotas $\mathcal Q$, initial $(q,m)$,
fixed throughput curve $G$, inner and outer thresholds $\epsilon_q,\epsilon_m$}
\For{each outer-loop update}{
  \For{each inner-loop update at fixed $m$}{
    Update the empirical work profile $\mathcal L$ with accepted complete-group token counts\;
    Observe $T_t^{\mathrm{tail}},T_t^{\mathrm{bulk}}$ and set
      $r_t^{\mathrm{tail}}\gets q/T_t^{\mathrm{tail}}$,
      $r_t^{\mathrm{bulk}}\gets(N-q)/T_t^{\mathrm{bulk}}$\;
    $\bar r_t^s\gets\beta\bar r_{t-1}^s+(1-\beta)r_t^s$,
      $s\in\{\mathrm{tail},\mathrm{bulk}\}$\;
    \lIf{mean normalized gap $\le\epsilon_q$}{\textbf{break}}
    $q_{\mathrm{balanced}}\gets\operatorname{clip}_{[\min\mathcal Q,\max\mathcal Q]}
      \!\left(N\bar r_t^{\mathrm{tail}}/(\bar r_t^{\mathrm{tail}}+\bar r_t^{\mathrm{bulk}})\right)$\;
    $q^+\gets\operatorname*{arg\,min}\nolimits_{q'\in
      \{\lfloor q_{\mathrm{balanced}}\rfloor,\lceil q_{\mathrm{balanced}}\rceil\}}
      \max\{q'/\bar r_t^{\mathrm{tail}},(N-q')/\bar r_t^{\mathrm{bulk}}\}$\;
    \lIf{$q^+\neq q$}{Apply $q^+$; restart the EMA and fresh stability evidence}
  }
  Predict $\widehat T^{\mathrm{tail}},\widehat T^{\mathrm{bulk}}$ from fixed $G$ and $\mathcal L$\;
  $\mathcal N(m)\gets(\{m\}\cup\{m-1,m+1\})\cap\{1,\ldots,R-1\}$\;
  $(m^+,q^+)\gets\operatorname*{arg\,min}\nolimits_{m'\in\mathcal N(m),\,q'\in\mathcal Q}
    \max\{\widehat T^{\mathrm{tail}}(q',m'),\widehat T^{\mathrm{bulk}}(q',m')\}$\;
  $\widehat T_{\mathrm{current}}\gets
    \min_{q'\in\mathcal Q}\max\{\widehat T^{\mathrm{tail}}(q',m),\widehat T^{\mathrm{bulk}}(q',m)\}$\;
  $\mathrm{gain}_{\mathrm{pred}}\gets
    (\widehat T_{\mathrm{current}}-\max\{\widehat T^{\mathrm{tail}}(q^+,m^+),
    \widehat T^{\mathrm{bulk}}(q^+,m^+)\})/\widehat T_{\mathrm{current}}$\;
  \lIf{$m^+=m$ or $\mathrm{gain}_{\mathrm{pred}}\le\epsilon_m$}{\Return $(q,m)$}
  $(m,q)\gets(m^+,q^+)$; reset the inner-loop state\;
}
\caption{\method's hierarchical allocation controller.}
\label{alg:tailsieve-controller}
\end{algorithm}

Increasing $q$ adds fresh tail groups while bulk still submits $N$ fresh
groups. Transition rounds do not update control measurements; quota decreases
reassign excess carry. Appendix~\ref{app:length-distribution} gives the
inventory and settlement conditions.

\subsection{Route-Specialized Speculative Decoding}
\label{sec:tail-speculation}

Tail isolation creates a low-concurrency stream with a long decoding
horizon. It permits deeper speculative proposals on the tail route,
while the bulk route uses a conservative depth to limit verification
cost. We instantiate this separation with native MTP and DFlash
\citep{qwen2026qwen35,chen2026dflash}; target-model verification preserves
the target distribution \citep{leviathan2023fast}. Routing and
speculation are complementary: routing controls the batches in which
verification executes, and each route can choose its backend and depth.
We also explored suffix decoding~\citep{oliaro2025suffixdecoding}, but limited accepted lengths and
ragged verification shapes that hinder full-batch CUDA-graph replay
outweighed its benefit; it is disabled by default.

\section{Evaluation}
\label{sec:evaluation}

\subsection{Experimental Overview}

We evaluate \method on five dense and mixture-of-experts model
configurations. These include Qwen3.5-35B-A3B, Qwen3.5-4B, and
Qwen3.5-2B~\citep{qwen2026qwen35}, together with
Qwen3-30B-A3B-Instruct-2507 and
Qwen3-4B-Instruct-2507~\citep{yang2025qwen3}.

All experiments run on the same eight-GPU server and use vLLM as the
rollout backend~\citep{kwon2023vllm}. Paired baseline and \method runs
use the same sampling configuration and required routing-unit count
\(N\).
The large-model configurations use four TP2 replicas, while the 4B and
2B configurations use eight TP1 replicas. The total replica budget
remains fixed, while \method jointly adjusts the routing-unit allocation
and the number of replicas assigned to the tail and bulk pools.
The step-wise experiments draw mathematical reasoning prompts from
DeepScaleR~\citep{agentica2025deepscaler} and coding prompts from the
KodCode-Light-RL-10K dataset on Hugging
Face~\citep{xu2025kodcode}; the end-to-end RL experiment uses only the
mathematical prompts. We use GRPO as the RL
algorithm~\citep{shao2024deepseekmath}. In our GRPO experiments, one
routing unit is a complete prompt group containing eight responses. We use a maximum output length of \(16\mathrm{K}\) tokens. Approximately \(3\%\) of responses are truncated on the math workload, compared with only \(0.1\%\) on the coding workload.
The shared hardware, sampling configuration, and prompt templates are
given in Appendix~\ref{app:common-experimental-setup}; the step-wise and
end-to-end protocols are detailed in
Appendices~\ref{app:step-wise-experiments}
and~\ref{app:end-to-end-rl-training}, respectively.

\subsection{Main Results}
\label{sec:eval-step}

We evaluate \method after the joint allocation \(q,m\) has converged and
the system has entered steady-state cross-round execution. We report
generation time at the selected settled operating point, averaged over
three runs with different random seeds. In each round, tail candidates
selected in the preceding round are routed to the tail pool, while the
current round produces the candidate set for the next round. Paired
baseline and routed runs share the prompt workload, sampling settings,
and total replica budget.

All selected requests are regenerated from scratch: no partial responses
or prefix tokens are reused, and all KV-cache state is flushed between
rounds. We use a rollout batch size of \(64\) groups, corresponding
to \(N=64\) routing units and \(512\) requests across all replicas.

We first disable speculative decoding and compare \method with
representative routing and scheduling strategies under the same
requests, model configuration, and total replica budget. Table~\ref{tab:routing-only} summarizes both routing-only workloads.

\begin{table}[!htbp]
\centering
\caption{Routing-only speedup over uniform group routing.}
\label{tab:routing-only}
\footnotesize
\setlength{\tabcolsep}{2pt}
\renewcommand{\arraystretch}{1.05}
\begin{tabularx}{\linewidth}{@{}L{0.34\linewidth}YYYYY@{}}
\toprule
\multicolumn{6}{c}{\textbf{DeepScaleR}} \\
\cmidrule(lr){1-6}
\multicolumn{1}{c}{Routing policy}
 & {\tiny\mbox{Qwen3.5-35B-A3B}}
 & {\tiny\mbox{Qwen3.5-4B}}
 & {\tiny\mbox{Qwen3.5-2B}}
 & {\tiny\mbox{Qwen3-30B-A3B}}
 & {\tiny\mbox{Qwen3-4B}} \\
\midrule
Uniform & 1.000$\times$ & 1.000$\times$ & 1.000$\times$ & 1.000$\times$ & 1.000$\times$ \\
Group LPT\textsuperscript{*}
 & 1.123$\times$ & 1.151$\times$ & 1.154$\times$ & \underline{1.085$\times$} & 1.225$\times$ \\
StreamRL-style Oracle\textsuperscript{*}
 & 1.030$\times$ & 1.113$\times$ & 1.136$\times$ & 0.928$\times$ & 1.174$\times$ \\
Seer-style Oracle\textsuperscript{*} & 1.189$\times$ & 1.162$\times$ & \underline{1.169$\times$} & 1.082$\times$ & \textbf{1.358$\times$} \\
\rowcolor{TailSieveLight}
\textsc{TailSieve} (Fixed Replica Allocation)
 & \underline{1.219$\times$} & \underline{1.199$\times$} & 1.144$\times$ & \textbf{1.112$\times$} & 1.186$\times$ \\
\rowcolor{TailSieveDeep}
\textbf{\textsc{TailSieve}}
 & \textbf{1.346$\times$} & \textbf{1.233$\times$} & \textbf{1.180$\times$} & \textbf{1.112$\times$} & \underline{1.254$\times$} \\
\bottomrule
\end{tabularx}

\vspace{10pt}

\begin{tabularx}{\linewidth}{@{}L{0.34\linewidth}YYYYY@{}}
\toprule
\multicolumn{6}{c}{\textbf{KodCode-10K}} \\
\cmidrule(lr){1-6}
\multicolumn{1}{c}{Routing policy}
 & {\tiny\mbox{Qwen3.5-35B-A3B}}
 & {\tiny\mbox{Qwen3.5-4B}}
 & {\tiny\mbox{Qwen3.5-2B}}
 & {\tiny\mbox{Qwen3-30B-A3B}}
 & {\tiny\mbox{Qwen3-4B}} \\
\midrule
Uniform & 1.000$\times$ & 1.000$\times$ & 1.000$\times$ & 1.000$\times$ & 1.000$\times$ \\
Group LPT\textsuperscript{*}
 & 1.033$\times$ & 1.001$\times$ & 1.106$\times$ & 1.083$\times$ & 1.107$\times$ \\
StreamRL-style Oracle\textsuperscript{*}
 & 1.167$\times$ & 1.033$\times$ & 1.100$\times$ & 1.094$\times$ & 0.992$\times$ \\
Seer-style Oracle\textsuperscript{*} & 1.282$\times$ & 0.971$\times$ & \underline{1.150$\times$} & 1.062$\times$ & \underline{1.406$\times$} \\
\rowcolor{TailSieveLight}
\textsc{TailSieve} (Fixed Replica Allocation)
 & \underline{1.560$\times$} & \underline{1.310$\times$} & 1.143$\times$ & \underline{1.166$\times$} & 1.179$\times$ \\
\rowcolor{TailSieveDeep}
\textbf{\textsc{TailSieve}}
 & \textbf{1.670$\times$} & \textbf{1.403$\times$} & \textbf{1.214$\times$} & \textbf{1.342$\times$} & \textbf{1.442$\times$} \\
\bottomrule
\end{tabularx}
\vspace{2pt}

\parbox{\linewidth}{\scriptsize \textsuperscript{*}Length-aware baselines use known historical group-average lengths.}
\end{table}

\begin{table}[!htbp]
\centering
\caption{Step-wise speedup over uniform routing with fixed replica
allocation using routing alone, MTP, or DFlash.}
\label{tab:step-speedup}
\footnotesize
\begin{tabular*}{\linewidth}{@{\extracolsep{\fill}}l*{6}{c}@{}}
\toprule
Model
& \multicolumn{3}{c}{DeepScaleR}
& \multicolumn{3}{c}{KodCode-10K} \\
\cmidrule(lr){2-4}\cmidrule(lr){5-7}
& Routing & +MTP & +DFlash & Routing & +MTP & +DFlash \\
\midrule
Qwen3.5-35B-A3B
 & 1.219$\times$ & 2.390$\times$ & 2.181$\times$ & 1.560$\times$ & 2.586$\times$ & 2.378$\times$ \\
Qwen3.5-4B
 & 1.199$\times$ & 2.414$\times$ & 2.091$\times$ & 1.310$\times$ & 2.408$\times$ & 2.148$\times$ \\
Qwen3.5-2B
 & 1.144$\times$ & 2.161$\times$ & -- & 1.143$\times$ & 2.349$\times$ & -- \\
\bottomrule
\end{tabular*}
\vspace{2pt}

\parbox{\linewidth}{\scriptsize \(-\) indicates that no official
DFlash implementation is available for Qwen3.5-2B.}
\end{table}

Uniform routing keeps each rollout group atomic. Group LPT uses each
group's average completion length from a matching historical run to
balance complete groups across replicas. The StreamRL-style and
Seer-style Oracle baselines use the same historical length estimates
and assign requests individually. These are routing-only adaptations
on the same vLLM executor, rather than reproductions of the full systems.
The fixed-allocation variant ablates \method's outer replica loop.
Appendix~\ref{app:step-wise-experiments} specifies the implementation of
each routing policy.

We next combine \method routing with speculative decoding. MTP uses
\(3\) speculative tokens, while DFlash uses a block size of \(4\).
The routing-with-speculation results form the second part of our main
comparison; Section~\ref{sec:eval-spec} further isolates the effect of
route-specific speculation, and Section~\ref{sec:eval-ablation}
examines decoding concurrency. Appendix~\ref{app:matched-speculation}
adds matched-speculation controls and absolute generation times, separating
the incremental routing gain from the combined gain over non-speculative
uniform routing.

\begin{figure*}[t]
    \centering
    \includegraphics[width=\textwidth]{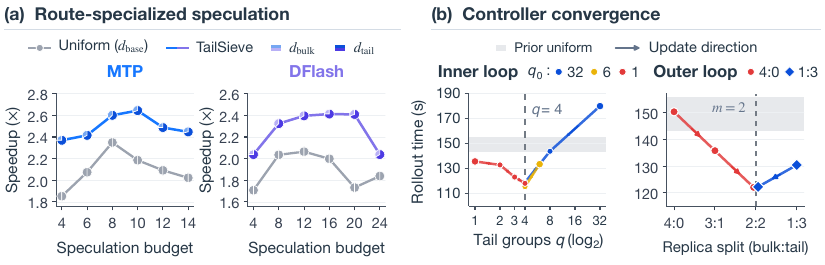}
    \caption{Route-specialized speculative decoding \textbf{(a)} and
    hierarchical-controller convergence \textbf{(b)} on
    Qwen3.5-35B-A3B.}
    \label{fig:spec-convergence}
\end{figure*}

\subsection{Route-Specialized Speculative Decoding}
\label{sec:eval-spec}

We further study how route specialization affects speculative decoding
on Qwen3.5-35B-A3B, as shown in
Figure~\ref{fig:spec-convergence}(a). For both MTP and DFlash, we compare
the two systems under the same total speculation budget. Let
\(d_{\mathrm{base}}\) denote the method-specific proposal depth used on
each baseline replica. The baseline uses the symmetric configuration
\((d_{\mathrm{base}},d_{\mathrm{base}})\), whereas \method may choose
route-specific depths satisfying
\[
d_{\mathrm{bulk}}+d_{\mathrm{tail}}
=2d_{\mathrm{base}}.
\]
For example, a baseline configuration of MTP-4/MTP-4 is compared with
MTP-3 on the high-concurrency bulk route and MTP-5 on the
low-concurrency tail route. This matched-budget comparison allows
\method to shift proposal depth from the less favorable bulk workload
to the tail workload, where more aggressive speculation remains
beneficial. The two methods use the same requests and sampling seeds.

Across the evaluated budgets, route-specialized speculation consistently
outperforms uniform routing. Tail isolation also shifts the best
speculation budget upward: the uniform baseline stops benefiting from
additional speculation earlier, whereas the isolated tail route remains
effective with a more aggressive budget. Thus, tail isolation both
increases the gain from speculative decoding and extends its useful
budget range.

\subsection{Convergence from Different Initial Allocations}
\label{sec:eval-convergence}

We evaluate both levels of the hierarchical controller while holding the
request stream and all other parameters fixed. For the inner loop, we
hold the replica split fixed and initialize the isolation size from the
minimum \((q_0=1)\), the empirical default \(q_0=6\), and the maximum
\((q_0=0.5N)\). Figure~\ref{fig:spec-convergence}(b) shows that all three
trajectories converge to \(q=4\), whether they approach it from above or
below. For the outer loop, we initialize the replica allocation from
uniform \((m_0=0)\) and a tail-heavy split \((m_0=3)\); both trajectories
converge to \(m=2\), corresponding to a \(2{:}2\) bulk--tail split. Thus,
both controller levels converge to the same allocation from substantially
different initial states. The points in Figure~\ref{fig:spec-convergence}(b)
report rollout time at settled configurations; they are not consecutive
online controller rounds. Appendix~\ref{app:controller-calibration}
adds the round-by-round inner-quota trajectories, with transition rounds
marked separately.

\subsection{End-to-End Rollout Generation during RL Training}
\label{sec:eval-e2e}

Step-wise acceleration is useful only if it persists as the policy
evolves. We therefore integrate \method into GRPO training and compare
it with uniform routing under the same initialization, prompt order,
number of policy updates, and sampling configuration. Our end-to-end
metric is the wall-clock time of the complete rollout generation stage
at each training step, including routing and controller execution. We
exclude the unchanged optimizer phase, which is outside the scope of
rollout routing. Appendix~\ref{app:end-to-end-rl-training} provides the
batch construction, optimizer settings, and random seed.

\suppressfloats[t]
\begin{figure*}[t]
    \centering
    \includegraphics[width=\textwidth]{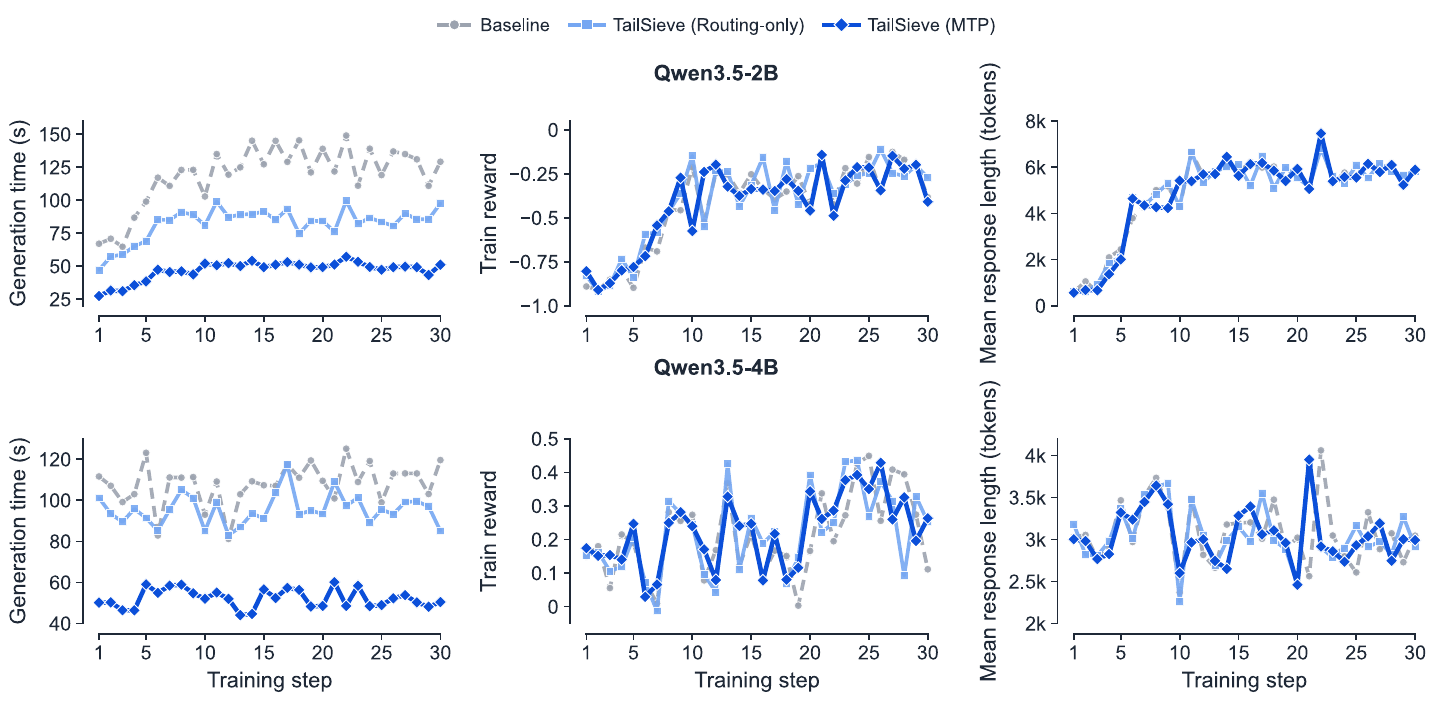}
    \caption{Thirty-step GRPO results for Qwen3.5-2B (top) and
    Qwen3.5-4B (bottom): generation time, reward, and mean response
    length.}
    \label{fig:rl-training}
\end{figure*}

Figure~\ref{fig:rl-training} shows that the generation-time improvement
persists as the policy evolves on both models. Qwen3.5-4B exhibits substantial step-to-step
variation in mean response length as the prompt batch changes. \method
tracks these workload shifts and continues to reduce generation time,
while MTP provides a larger and more consistent reduction. Despite the
natural evolution of response length during training, the reward and
mean-response-length curves remain comparable across the three runs on
both models. Thus, the observed acceleration does not come with a
systematic shift in either training reward or average response length.

\begin{table}[t]
\centering
\caption{Downstream accuracy (AIME: mean over three evaluation seeds);
parentheses report percentage-point changes from the corresponding base model.}
\label{tab:rl-accuracy}
\scriptsize
\setlength{\tabcolsep}{2pt}
\renewcommand{\arraystretch}{0.9}
\begin{tabularx}{\linewidth}{@{}>{\centering\arraybackslash}m{0.12\linewidth}>{\raggedright\arraybackslash}m{0.22\linewidth}YYYY@{}}
\toprule
Model & Method
& GSM8K pass@1
& MATH-500 pass@1
& AIME24 pass@10
& AIME25 pass@10 \\
\midrule
\multirow{4}{*}{Qwen3.5-2B}
& Base model
& \(9.40\%\)
& \(13.40\%\)
& \(33.33\%\)
& \(26.67\%\) \\
& Baseline
& \mbox{\(75.97\%\) \((+66.57)\)}
& \mbox{\(76.20\%\) \((+62.80)\)}
& \mbox{\(56.67\%\) \((+23.33)\)}
& \mbox{\(36.67\%\) \((+10.00)\)} \\
& \method (Routing-only)
& \mbox{\(76.57\%\) \((+67.17)\)}
& \mbox{\(74.60\%\) \((+61.20)\)}
& \mbox{\(57.78\%\) \((+24.44)\)}
& \mbox{\(41.11\%\) \((+14.44)\)} \\
& \method (MTP)
& \mbox{\(76.42\%\) \((+67.02)\)}
& \mbox{\(74.40\%\) \((+61.00)\)}
& \mbox{\(53.33\%\) \((+20.00)\)}
& \mbox{\(36.67\%\) \((+10.00)\)} \\
\midrule
\multirow{4}{*}{Qwen3.5-4B}
& Base model
& \(89.61\%\)
& \(84.40\%\)
& \(90.00\%\)
& \(82.22\%\) \\
& Baseline
& \mbox{\(89.16\%\) \((-0.45)\)}
& \mbox{\(88.80\%\) \((+4.40)\)}
& \mbox{\(91.11\%\) \((+1.11)\)}
& \mbox{\(80.00\%\) \((-2.22)\)} \\
& \method (Routing-only)
& \mbox{\(89.46\%\) \((-0.15)\)}
& \mbox{\(87.40\%\) \((+3.00)\)}
& \mbox{\(91.11\%\) \((+1.11)\)}
& \mbox{\(82.22\%\) \((+0.00)\)} \\
& \method (MTP)
& \mbox{\(89.61\%\) \((+0.00)\)}
& \mbox{\(86.60\%\) \((+2.20)\)}
& \mbox{\(90.00\%\) \((+0.00)\)}
& \mbox{\(84.44\%\) \((+2.22)\)} \\
\bottomrule
\end{tabularx}
\end{table}

Table~\ref{tab:rl-accuracy} shows that the final checkpoints from both
\method variants retain broadly comparable downstream accuracy to the
baseline across all four benchmarks. We evaluate checkpoint quality on
GSM8K and MATH-500~\citep{cobbe2021training,lightman2024lets}, and the
2024 and 2025 AIME problem sets~\citep{maa2026aime}, reporting pass@1
for the former and pass@10 for the latter. AIME scores are means over
three evaluation sampling seeds at each fixed checkpoint;
Appendix~\ref{app:downstream-uncertainty} gives the evaluation protocol,
per-seed scores, and standard deviations.

\subsection{Scaling with Concurrency}
\label{sec:eval-ablation}

To isolate the effect of decoding concurrency from speculative decoding,
we disable speculation and vary the number of rollout groups. At each
concurrency level, the baseline and routing configurations use the same
randomly sampled requests, sampling seeds, and generation settings.
Table~\ref{tab:concurrency} shows the resulting wall-clock speedups.
Here, \(N\) denotes the number of GRPO prompt groups, with eight
responses per group.

\begin{table}[t]
\centering
\caption{Routing-only speedup over uniform routing as rollout
concurrency increases.}
\label{tab:concurrency}
\footnotesize
\begin{tabular*}{\linewidth}{@{\extracolsep{\fill}}l*{6}{c}@{}}
\toprule
Model
& \multicolumn{3}{c}{DeepScaleR}
& \multicolumn{3}{c}{KodCode-10K} \\
\cmidrule(lr){2-4}\cmidrule(lr){5-7}
& N=64 & N=128 & N=256
& N=64 & N=128 & N=256 \\
\midrule
Qwen3.5-35B-A3B
 & 1.346$\times$ & 1.199$\times$ & 1.122$\times$
 & 1.670$\times$ & 1.214$\times$ & 1.142$\times$ \\
Qwen3.5-4B
 & 1.233$\times$ & 1.225$\times$ & 1.191$\times$
 & 1.403$\times$ & 1.453$\times$ & 1.240$\times$ \\
Qwen3.5-2B
 & 1.180$\times$ & 1.129$\times$ & 1.177$\times$
 & 1.214$\times$ & 1.108$\times$ & 1.101$\times$ \\
Qwen3-30B-A3B-Instruct-2507
 & 1.112$\times$ & 1.023$\times$ & 1.121$\times$
 & 1.342$\times$ & 1.163$\times$ & 1.091$\times$ \\
Qwen3-4B-Instruct-2507
 & 1.254$\times$ & 1.215$\times$ & 1.306$\times$
 & 1.442$\times$ & 1.367$\times$ & 1.248$\times$ \\
\bottomrule
\end{tabular*}
\end{table}

\section{Related Work}

\subsection{Long-Tail RL Rollouts and Partial Rollout}

Long responses create synchronization bubbles in LLM RL. AReaL relaxes
this barrier by decoupling rollout generation from training
\citep{fu2025areal}. Kimi k1.5 reuses segments of previous trajectories
\citep{kimi2025k15}, while APRIL and CoPRIS overprovision rollouts and carry
unfinished trajectories into later steps, with CoPRIS correcting cross-policy
reuse through importance sampling \citep{zhouapril2025,qu2025copris}.
RollPacker instead consolidates tail prompts into a small number of long
rounds; its short rounds also generate extra responses per prompt and
retain the first ones to finish \citep{gao2026rollpacker}.

\method uses cutoff status only to select prompt groups for fresh
generation under the current policy, mixing tail and bulk groups within
each synchronous update.

\subsection{Length-Aware Scheduling and Routing}

Serving schedulers use proxy models, uncertainty-aware distributions, or
entropy-guided representations to predict response lengths
\citep{qiu2024ssjf,zheng2026tie,xie2026forelen}. In RL systems, StreamRL uses
a learned output-length ranker for skewness-aware dispatching
\citep{zhong2025streamrl}, while Seer exploits similarities among responses to
the same prompt through online context learning and divided rollout
\citep{qin2026seer}. Libra~\citep{chen2026libra} uses tool-return signals to estimate residual
trajectory lengths, routes ongoing requests among heterogeneous
tensor-parallel buckets, and reallocates workers between rollout and
training. \method instead uses partial-rollout status for prompt-group
selection and jointly adapts tail quota and replica capacity within a
fixed rollout budget, without an auxiliary length predictor.

\subsection{Speculative Rollout Decoding}

Speculative rollout methods reduce generation cost using adaptive drafts,
history from nearby rollouts, idle-compute pre-generation, or improved
multi-token prediction
\citep{shao2026das,he2026rhymerl,liu2025specrl,xububblespec2026,li2026bebop}.
These techniques accelerate token generation, whereas \method shapes the
concurrency at which generation runs. The two dimensions are complementary:
tail isolation creates a low-concurrency, long-horizon route on which deeper
MTP or DFlash speculation can be applied without imposing the same policy on
the high-concurrency bulk route.

\section{Conclusion}

We presented \method, a partial-rollout-guided framework for mitigating long-tail stalls in synchronous LLM rollouts. Our key insight is that isolating a small amount of tail workload can reduce step makespan, but the optimal degree of isolation depends on the evolving workload and decoding concurrency. \method turns this insight into an online system by using partial rollouts to identify tail-prone prompt groups and jointly adapting tail workload and replica capacity. Across evaluated models and workloads, \method achieves up to \(1.67\times\) routing-only speedup and \(2.59\times\) combined speedup with routing and route-specialized speculative decoding, while maintaining comparable training quality. The gains are strongest when long-tail generations determine rollout
makespan and tail isolation provides a decoding-concurrency advantage.

\clearpage
\section*{AI use statement}
Generative AI tools were used to assist with language editing, improve
clarity, and polish the presentation of this manuscript.
\section*{Reproducibility statement}
Appendix~\ref{app:controller-calibration} gives the executable controller
rules and parameters, Appendix~\ref{app:experimental-details} records
experimental configurations, and Appendix~\ref{app:oracle-proof}
specifies the offline oracle and its exactness argument.

\bibliographystyle{iclr2026_conference}
\bibliography{iclr2026_conference}

\clearpage
\appendix
\numberwithin{table}{section}
\numberwithin{figure}{section}

\begin{center}
    {\Large\bfseries Appendix}
\end{center}

\section{Additional Evaluation Details}
\label{app:experimental-details}

\subsection{Common Experimental Setup}
\label{app:common-experimental-setup}

Experiments run on one node with
eight NVIDIA H100 GPUs and use vLLM as the rollout
backend~\citep{kwon2023vllm}. Experiments use
vLLM \texttt{0.23.1rc1.dev562+g3483240b7}, based on main commit
\href{https://github.com/vllm-project/vllm/commit/3483240b7ea3d4372b6c79369ea36617f8b1fbb2}{\texttt{3483240b7}} with local
Qwen3.5/V2 compatibility changes. These changes cover LM-head access,
attention metadata for the draft and target models, and hybrid cache handling.
Its Model Runner V2 (MRV2) DFlash path
derives from
\href{https://github.com/vllm-project/vllm/pull/44586}{\texttt{vllm-project/vllm} PR~\#44586},
including full CUDA-graph support for DFlash. The two large-model
configurations use four TP2 replicas, while the 4B and 2B configurations
use eight TP1 replicas. The total replica budget remains fixed in every
comparison.

The step-wise experiments use mathematical prompts from
DeepScaleR~\citep{agentica2025deepscaler} and bounded-reasoning coding
prompts from \texttt{KodCode/KodCode-Light-RL-10K}~\citep{xu2025kodcode}.
Step-wise results are averaged over three runs with different random seeds.
The RL experiments use the mathematical workload. Responses are sampled
independently in each run, so realized response lengths and token counts
may differ across otherwise matched configurations.
Table~\ref{tab:shared-sampling-config} reports the sampling settings shared
by step-wise evaluation and RL training rollouts.

\begin{table}[!htbp]
\centering
\caption{Rollout sampling configuration shared by the step-wise and
end-to-end RL experiments.}
\label{tab:shared-sampling-config}
\small
\begin{tabularx}{0.76\linewidth}{@{}C{0.34\linewidth}>{\centering\arraybackslash}X@{}}
\toprule
Parameter & Setting \\
\midrule
Maximum prompt length & \(512\) tokens \\
Maximum response length & \(16{,}384\) tokens \\
Temperature & \(0.7\) \\
Top-\(p\) & \(0.8\) \\
Top-\(k\) & \(20\) \\
Presence penalty & \(1.5\) \\
Thinking mode & Disabled \\
Stop sequence & \texttt{<END>} \\
\bottomrule
\end{tabularx}
\end{table}

\paragraph{Prompt templates.}
The DeepScaleR workload uses no system message and applies the following
user template, where \texttt{\{question\}} is the dataset problem:
\begin{promptbox}{Mathematics --- User}
\{question\}\\
Please solve the problem step by step. End your response with exactly one final answer in this form:\\
\textbackslash boxed\{...\}<END>\\
After <END>, stop immediately. Do not write any words,
punctuation, newline, or explanation after \texttt{<END>}.
\end{promptbox}

The bounded-reasoning code workload is used only in the step-wise
routing and MTP experiments. Its messages are:
\begin{promptbox}{Code --- System}
You are an expert Python programmer.
\end{promptbox}
\begin{promptbox}{Code --- User}
Solve the coding problem below. Provide your reasoning and explanation,
including the key algorithm, its correctness, important edge cases, and
its time and space complexity. Then provide exactly one complete Python
function. After the function, write <END> and stop immediately.
Do not repeat or revise the answer.

\medskip
Problem:\\
\{question\}

\medskip
Required function:\\
\{function\_declaration\}

\medskip
Specification:\\
\{docstring\}
\end{promptbox}

\subsection{Step-Wise Rollout Experiments}
\label{app:step-wise-experiments}

\paragraph{Routing-policy implementations.}
Uniform group routing distributes complete groups evenly by count for
the divisible group counts used in the baseline comparisons. All marked
length-aware baselines in Table~\ref{tab:routing-only} use each prompt
group's average completion length from a matching historical run.

\emph{Group LPT (historical-length group scheduling).}
Group~LPT (longest processing time first) sorts whole eight-response
groups by decreasing historical mean length and greedily assigns each
group to the replica with the lowest accumulated predicted load among
those below the equal
group-count capacity.

\emph{StreamRL-style Oracle (historical-length skewness-aware routing).}
This baseline adapts skewness-aware dispatching~\citep{zhong2025streamrl}
at request granularity. Each response inherits its group's historical
mean length. We sort the $n=512$ requests in decreasing order of this
estimate, assign the first $\lfloor 0.2n\rfloor=102$ to a tail pool,
and place the remainder in
a regular pool. We enumerate all nonempty replica splits and select the
one minimizing the sum of the two pools' predicted latencies. The
prediction uses an offline per-token latency profile, KV-cache capacity,
and batch-size limits, with the 90th and 50th percentiles of the
historical length distribution representing the tail and regular pools,
respectively. Within each pool, sorted requests are distributed
round-robin across replicas and admitted in decreasing predicted-length
order on each replica. This allocation can split a prompt group across
replicas or across the two pools.

\emph{Seer-style Oracle (historical-length request LPT).}
This baseline uses request-level LPT placement inspired by
Seer~\citep{qin2026seer}. Each response is weighted by its
group's historical mean length. Requests are sorted by decreasing weight
and greedily assigned to the replica with the lowest accumulated
predicted load among those below the equal request-count capacity
$n/R$. Responses from the same prompt group may be placed on different
replicas.

The \method executor uses group-sticky placement for the evenly divisible
prefix of groups in each pool and request-level scheduling for remainder
groups. Group-level selection and completion do not imply that every
response of every group is physically assigned to one replica.

The fixed-allocation \method variant uses the same partial-rollout
selector and inner cutoff feedback, but holds
the replica split at $2{:}2$ for four-TP2-replica configurations and
$4{:}4$ for eight-TP1-replica configurations. Full \method additionally
adjusts the replica split through its outer loop.

All timings report rollout generation at a settled operating point. The
unfinished-group partition initializes the following round, but generation
restarts from the original prompt: generated prefixes and cross-round KV
state are not reused.

The route-specialization sweep compares symmetric and route-specific proposal
depths under the matched depth budget defined in
Appendix~\ref{app:speculative-decoding-setup}.

\paragraph{Rollout-concurrency protocol.}
To isolate the effect of decoding concurrency from speculative decoding,
we disable speculation and vary the number of rollout groups under a fixed
replica budget. At each concurrency level, the baseline and routing
configurations use the same sampled requests, sampling seeds, and generation
settings. Table~\ref{tab:concurrency} reports the routing-only speedup, where
$N$ is the number of prompt groups and each group contains eight responses.
For $N=128$ and $N=256$, all experimental settings are kept identical to
those at $N=64$, except for the number of rollout groups.
See Appendix~\ref{app:conditional-gain} for a discussion of the gain regimes.

\subsection{Speculative-Decoding Configurations and Depth Selection}
\label{app:speculative-decoding-setup}

With route-specific speculative configurations, changing replica roles can
require runtime reconfiguration or additional CUDA-graph captures and
persistent buffers. To avoid these switching and memory costs, we evaluate
speculative decoding with fixed replica pools while adapting the tail quota.

For equal-sized replica pools, matching the replica-weighted proposal-depth
budget reduces to $d_{\mathrm{bulk}}+d_{\mathrm{tail}}=2d_{\mathrm{base}}$,
as used in Figure~\ref{fig:spec-convergence}(a). For a general split with
$R$ replicas and $m$ tail replicas, the corresponding condition is
$(R-m)d_{\mathrm{bulk}}+m d_{\mathrm{tail}}=R d_{\mathrm{base}}$.
This budget measures configured proposal depth; actual drafting and
verification FLOPs and runtime also depend on the workload and acceptance
behavior.

For Figure~\ref{fig:spec-convergence}(a), we select route-specific proposal
depths through offline profiling on the same 64-prompt workload. At each
matched depth budget, we choose the tested pair with the lowest mean
generation wall time over settled rounds at the final quota. The selected
depths remain fixed at each run.

\subsection{End-to-End Mixed-Routing RL Training}
\label{app:end-to-end-rl-training}

The end-to-end experiment trains Qwen3.5-2B and Qwen3.5-4B. To maintain training
stability, we follow Dr.\ GRPO~\citep{liu2025understanding} for advantage
normalization:
advantages are mean-centered within each eight-response prompt group
but are not divided by the within-group reward standard deviation.
The RL warm-up runs for one step with the same batch size as training,
and its responses are routed across all replicas. Its batch is reserved
from the end of the dataloader and excluded from training, so warm-up
does not alter the data consumed by either the baseline or routing run.
The reward includes a DAPO-style linear overlength
penalty~\citep{yu2025dapo}. It is zero up to
\(8{,}192\) tokens and, for a response of length \(L>8{,}192\), reduces
the reward by \((L-8{,}192)/8{,}192\). This corresponds to a reduction of
\(0.125\) per additional \(1{,}024\) tokens and reaches \(1.0\) at the
\(16{,}384\)-token limit. Table~\ref{tab:rl-training-config} reports the
batch construction, optimizer settings, and random seed.

\begin{table}[!htbp]
\centering
\caption{Batch and optimizer configuration for the end-to-end
mixed-routing RL experiment.}
\label{tab:rl-training-config}
\small
\begin{tabularx}{0.76\linewidth}{@{}C{0.34\linewidth}>{\centering\arraybackslash}X@{}}
\toprule
Parameter & Setting \\
\midrule
Batch size & \(64\) groups \(\times\,8\) responses \(=512\) trajectories \\
Training TP & \(1\) \\
Training micro-batch size & \(8\) \\
PPO clip ratio & \(0.2\) \\
Dual-clip coefficient & \(3.0\) \\
KL loss/reward & Disabled \\
Entropy coefficient & \(0\) \\
Learning rate & \(2\times10^{-7}\) \\
Learning-rate schedule & Constant \\
Adam betas & \((0.9,0.999)\) \\
Weight decay & \(0.01\) \\
Gradient clipping & \(1.0\) \\
Random seed & \(42\) \\
\bottomrule
\end{tabularx}
\end{table}

\subsection{Exact Offline Routing Oracle}
\label{app:oracle-proof}

This subsection specifies an exact oracle for the two-replica assignment
problem used in our motivating analysis. Exactness refers to optimization
under a fixed all-admit decoding model; it does not imply an exact model
of a serving runtime or an optimal online controller.

\paragraph{Model and scope.}
There are $n$ requests with known completion lengths
$L_1\ge\cdots\ge L_n\ge0$.
The two replicas are identical. All requests assigned to a replica begin
decoding together, advance by one token per decoding iteration, and leave
upon completion. The duration of an iteration with $b$ active requests is
the fixed, finite value $\tau(b)>0$ for $1\le b\le n$; set $\tau(0)=0$.
Thus, iteration time depends only on active request count. We exclude prefill, admission queues, memory-capacity
constraints, and replica interference from this oracle model. An empty
replica has completion time zero. The objective is
\begin{equation}
  T^\star=\min_{S\subseteq\{1,\ldots,n\}}
  \max\{T(S),T(S^{\mathrm c})\},
  \label{eq:oracle-objective}
\end{equation}
where $T$ is the replica time in Equation~\ref{eq:decoding-time}.

\paragraph{Sorted insertion cost.}
Suppose a replica already holds $c$ requests from a prefix of the sorted
list, and the next request has length $L_i$. Each existing length is at
least $L_i$. For the first $L_i$ decoding positions, adding the request
changes the active count from $c$ to $c+1$; later positions are unchanged.
Consequently, its incremental completion time is
\begin{equation}
  d_i(c)=L_i\bigl[\tau(c+1)-\tau(c)\bigr].
  \label{eq:oracle-increment}
\end{equation}
The formula includes $c=0$.

\paragraph{State and recurrence.}
After assigning the first $i$ requests, let $\mathcal F_{i,c}$ contain
the nondominated attainable completion-time pairs $(u,v)$ for assignments
with exactly $c$ requests on replica A and $i-c$ on replica B.
A pair $(u,v)$ dominates $(u',v')$ if $u\le u'$ and $v\le v'$, with at
least one strict inequality; duplicate pairs need only be stored once.
Write $\operatorname{ND}$ for removal of dominated pairs and duplicates.
Initialize $\mathcal F_{0,0}=\{(0,0)\}$, with all invalid states empty.
For $i=0,\ldots,n-1$ and $c=0,\ldots,i+1$, the recurrence is
\begin{equation}
\begin{aligned}
\mathcal C^{\mathrm A}_{i+1,c}
  &=\{(u+d_{i+1}(c-1),v):(u,v)\in\mathcal F_{i,c-1}\},\\
\mathcal C^{\mathrm B}_{i+1,c}
  &=\{(u,v+d_{i+1}(i-c)):(u,v)\in\mathcal F_{i,c}\},\\
\mathcal F_{i+1,c}
  &=\operatorname{ND}\bigl(
       \mathcal C^{\mathrm A}_{i+1,c}\cup
       \mathcal C^{\mathrm B}_{i+1,c}\bigr).
\end{aligned}
\label{eq:oracle-recurrence}
\end{equation}
The first set is included only when $1\le c\le i+1$, and the second only
when $0\le c\le i$, so $d_{i+1}$ is never evaluated at a negative count.
The oracle returns
\begin{equation}
  \min_{0\le c\le n}\ \min_{(u,v)\in\mathcal F_{n,c}}\max\{u,v\}.
  \label{eq:oracle-terminal}
\end{equation}
Storing a predecessor and the replica chosen for each retained pair
recovers a corresponding assignment. Different values of $c$ must not
be pruned against one another, since their future insertion costs can
differ.

\paragraph{Exactness.}
Without pruning, the two transitions cover every assignment using the exact
insertion costs in Equation~\ref{eq:oracle-increment}. For candidates in
the same state $(i,c)$, any fixed assignment of the remaining requests
adds the same pair of future cost increments to each candidate. A dominated pair
therefore cannot yield a smaller final makespan than its dominating pair
under any continuation. With exact arithmetic and comparisons, pruning
preserves an optimal completion, so the terminal minimization in
Equation~\ref{eq:oracle-terminal} returns $T^\star$.

\subsection{Offline Routing Simulation Across Rollout Groups}
\label{app:offline-routing-simulation}

\begin{figure}[!htbp]
    \centering
    \includegraphics[width=0.58\linewidth]{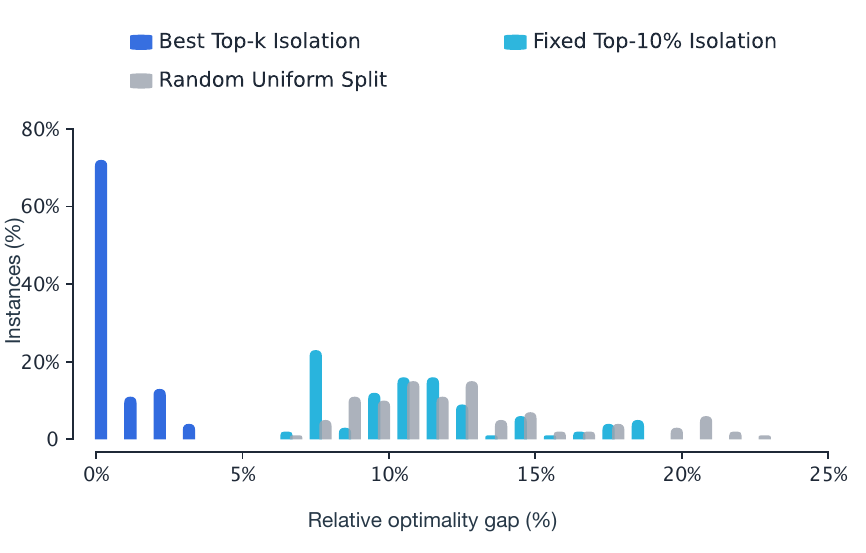}
    \caption{Optimality-gap distributions across 100 rollout groups.}
    \label{fig:appendix-tail-isolation}
\end{figure}

We repeat the offline routing comparison on 100 sampled rollout groups.
Each group contains 100 trajectories whose lengths are sampled from
real rollout traces. The simulator's GPU decoding characteristics are
calibrated using direct measurements from Qwen3.5-35B-A3B. For a routing
policy with makespan \(T_{\mathrm{policy}}\), we measure its relative
optimality gap as
\[
\frac{T_{\mathrm{policy}}-T_{\mathrm{opt}}}{T_{\mathrm{opt}}},
\]
where \(T_{\mathrm{opt}}\) is the exact offline optimum computed by the
dynamic program. The per-group best-\(k\) isolation policy is within
\(4\%\) of the exact optimum for all 100 groups and within \(1\%\) for
\(72\%\) of them. In comparison, fixed \(10\%\) isolation and random
uniform routing produce broader, right-shifted gap distributions, as
shown in Figure~\ref{fig:appendix-tail-isolation}.

\subsection{Tail-Strength Workload Construction}
\label{app:tail-strength-workloads}

To vary tail strength without changing request identity or ordering, we
derive the weak- and strong-tail workloads from the same measured
response-length trace used by the real workload. Let \(M\) and
\(P_{90}\) denote the median and \(90\)th-percentile length of the real
trace, respectively. For a target ratio \(r\), we transform only lengths
above the median:
\[
L_i'(r)=
\begin{cases}
L_i, & L_i\le M,\\
M+s_r(L_i-M), & L_i>M,
\end{cases}
\qquad
s_r=\frac{rM-M}{P_{90}-M}.
\]
The weak- and strong-tail workloads use \(r=1.5\) and \(r=6\),
respectively, while the real workload directly uses
the original lengths. Because \(s_r>0\), the transformation preserves the
ordering of requests. It also leaves the lower half and median unchanged
and maps the original \(P_{90}\) to \(rM\), making
\(P_{90}/P_{50}=r\). Thus, the three workloads differ only in the scale
of the upper tail. Table~\ref{tab:tail-strength-workloads} summarizes
their resulting length statistics.

\begin{table}[!htbp]
\centering
\caption{Construction and length statistics of the weak-, real-, and
strong-tail workloads.}
\label{tab:tail-strength-workloads}
\small
\begin{tabularx}{0.96\linewidth}{@{}C{0.15\linewidth}Y C{0.09\linewidth}C{0.09\linewidth}C{0.12\linewidth}C{0.10\linewidth}@{}}
\toprule
Scene & Length construction & \(P_{50}\) & \(P_{90}\) & \(P_{90}/P_{50}\) & Maximum \\
\midrule
Weak Tail & Compress the upper half & \(1{,}160\) & \(1{,}740\) & \(1.5\) & \(3{,}206\) \\
Real Workload & Original lengths & \(1{,}160\) & \(4{,}115\) & \(3.547\) & \(11{,}800\) \\
Strong Tail & Stretch the upper half & \(1{,}160\) & \(6{,}960\) & \(6.0\) & \(22{,}094\) \\
\bottomrule
\end{tabularx}
\end{table}

\section{Long-Tail Prompts across Policy Updates}
\label{app:prompt-stability}

\subsection{Partial-Rollout Cutoff and Coverage}
\label{app:partial-cutoff}

\paragraph{Experimental setup.}
This study uses Qwen3.5-35B-A3B on DeepScaleR with four TP2 vLLM
replicas and speculative decoding disabled. Each replica permits at most
256 running sequences. The study follows the mathematics
prompt template and sampling settings in
Appendix~\ref{app:common-experimental-setup} and
Table~\ref{tab:shared-sampling-config}, with repetition penalty $1.0$
and a maximum context length of $16{,}896$ tokens.

Figure~\ref{fig:partial-filter} uses prompt groups, each containing eight
responses. A group finishes when all eight responses complete; its final
length is the maximum number of generated tokens among those responses.
The cutoff study uses ten fully completed rollout traces, each containing
250 groups, with a completion quota of $M=200$ groups.

\paragraph{Cutoff definition.}
For a cutoff parameter $\rho$, let $k=\rho M$. Within each trace, we take
the first $M+k$ groups in submission order. The virtual cutoff is the
recorded time at which the $M$-th group in this subset finishes; the
remaining $k$ groups form the candidate set $P$. This gives
\[
\rho=\frac{\text{excess groups}}{\text{completion quota}}
=\frac{k}{M},
\qquad
M+k=(1+\rho)M.
\]
Thus $\rho=5\%,10\%,25\%$ corresponds to subsets of 210, 220, and
250 groups, leaving 10, 20, and 50 candidates, respectively. The cutoffs
are evaluated retrospectively using completion times from the full
250-group traces, so final lengths are available for every candidate.

\paragraph{Coverage and example.}
Within the same subset, let $O$ be the $k$ groups with the largest final
group lengths. Coverage is the fraction of this length-ranked tail set
captured by the unfinished candidates,
\[
\operatorname{coverage}(\rho)
=\frac{1}{10}\sum_{s=1}^{10}
\frac{|P_s\cap O_s|}{k}.
\]
For example, $\rho=10\%$ adds 20 groups to the 200-group quota.
The virtual cutoff leaves 20 of the 220 groups unfinished; these are
compared with the 20 longest groups in that same subset. Across ten
traces, 186 of the 200 length-ranked tail groups are selected, giving
$186/200=93\%$ coverage. The corresponding totals at $5\%$ and $25\%$
are $85/100=85\%$ and $474/500=94.8\%$. Figure~\ref{fig:partial-filter}(c)
rounds these values to whole percentages: $85\%$, $93\%$, and $95\%$.

\paragraph{Cutoff--length mismatch.}
The cutoff selection and final-length ranking use the same generated
responses within each trace. Let $f_{g,r}$ be response $r$'s completion
time measured from the start of the rollout, and let $\ell_{g,r}$ be its
generated-token count. The two group statistics are
\[
F_g=\max_{r=1,\ldots,8}f_{g,r},
\qquad
L_g=\max_{r=1,\ldots,8}\ell_{g,r}.
\]
The candidate set $P$ contains the groups with the latest $F_g$, whereas
$O$ contains those with the largest $L_g$. The serving engine caps the
number of running sequences, so responses are admitted at different
times as capacity becomes available. Completion times therefore include
admission waiting: an earlier-admitted long response can finish before
a later-admitted shorter response. Completion order therefore depends
on both admission time and generation duration. The yellow
uncovered groups in Figure~\ref{fig:partial-filter}(b) represent $O\setminus P$:
they belong to the length-ranked tail but have already finished at the
virtual cutoff. Thus the mismatch measures the difference between
wall-clock completion and token-length rankings within the same rollout.

\subsection{Cross-Checkpoint Tail Ranking and Length Variation}

We analyze whether long-tail behavior is driven primarily by sampling
randomness, policy changes, or the prompt. We do not assume that absolute
response lengths or the marginal length distribution remain fixed
during training. Instead, we ask whether prompt identity remains the
dominant source of cross-request length heterogeneity. Let
\(L_{i,s,r}\) denote the raw response length for prompt \(i\), checkpoint
\(s\), and sampling repeat \(r\). We use \(I=64\) prompts, \(S\)
checkpoints, and \(R=8\) samples per prompt and checkpoint, without
normalizing the lengths.

\begin{figure*}[t]
    \centering
    \includegraphics[width=\textwidth]{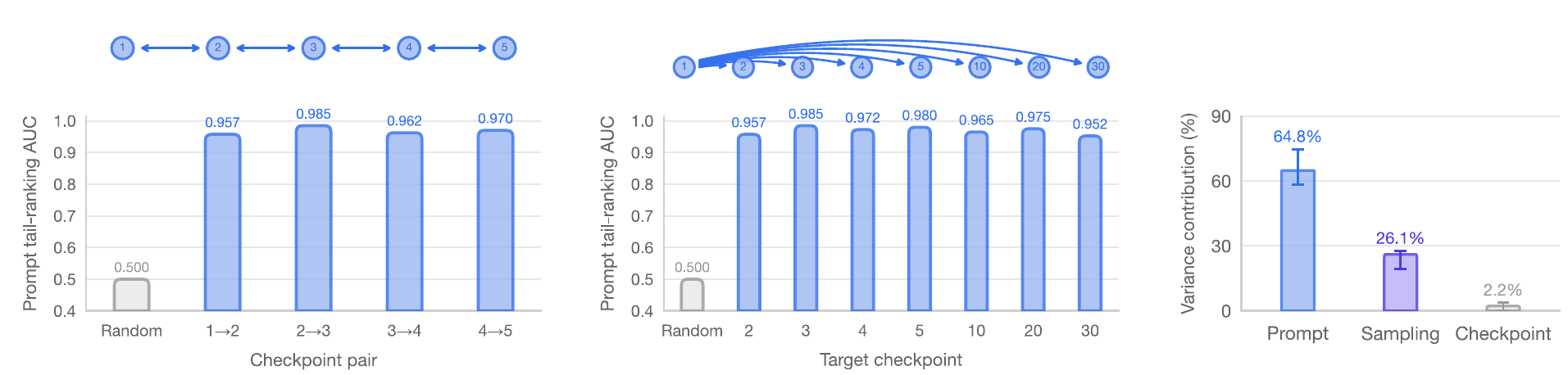}
    \caption{Qwen3.5-2B prompt-tail ranking stability across checkpoints and
    response-length variance decomposition.}
    \label{fig:prompt-tail-auc}
\end{figure*}

\paragraph{Cross-checkpoint tail ranking.}
For each prompt and checkpoint, we first average its eight sampled
lengths,
\[
\bar L_{i,s}=\frac{1}{R}\sum_{r=1}^{R}L_{i,s,r}.
\]
For a source checkpoint \(s\), we define its tail set \(\mathcal T_s\)
as the \(\lceil0.1I\rceil=7\) prompts with the largest \(\bar L_{i,s}\).
At a target checkpoint \(t\), we use \(\bar L_{i,t}\) as the ranking
score and compute
\[
\mathrm{AUC}(s,t)
=
\Pr\!\left(
\bar L_{i,t}>\bar L_{j,t}
\mid i\in\mathcal T_s,\;j\notin\mathcal T_s
\right).
\]
Figure~\ref{fig:prompt-tail-auc} reports \(0.957\)--\(0.985\) for
adjacent checkpoints and \(0.952\)--\(0.985\) when checkpoint 1 ranks
checkpoints through 30, well above the random baseline of \(0.5\).

\paragraph{Variance contribution.}
\label{app:length-variance}
We further apply a two-factor decomposition directly to
\(L_{i,s,r}\):
\[
SS_{\mathrm{total}}
=SS_{\mathrm{prompt}}+SS_{\mathrm{checkpoint}}
+SS_{\mathrm{interaction}}+SS_{\mathrm{sampling}}.
\]
Using dots to denote averages over the corresponding indices, the main
terms are
\[
SS_{\mathrm{prompt}}
=SR\sum_i(\bar L_{i..}-\bar L_{...})^2,
\qquad
SS_{\mathrm{checkpoint}}
=IR\sum_s(\bar L_{.s.}-\bar L_{...})^2,
\]
and
\[
SS_{\mathrm{sampling}}
=\sum_{i,s,r}(L_{i,s,r}-\bar L_{is.})^2,
\]
with the interaction given by the remaining between-cell variation.
Each contribution is normalized as
\[
p_f=\frac{SS_f}{SS_{\mathrm{total}}}\times100\%.
\]
For Qwen3.5-2B, prompt identity contributes \(64.8\%\), compared with
\(26.1\%\) from sampling, \(2.2\%\) from checkpoints, and
\(6.9\%\) from prompt--checkpoint interaction. The additional
Qwen3.5-4B decomposition gives \(77.78\%\), \(20.26\%\),
\(0.01\%\), and \(1.95\%\), respectively. These are
model-specific contributions, rather than an average across models.

These results do not imply that sampling randomness has no effect, that
the overall response-length distribution is stationary, or that an
individual prompt can never move into or out of the tail. Instead, they
support the simpler intuition used by \method: long-tail prompts tend to
remain long-tailed across adjacent policy updates because differences
between prompts dominate the variation introduced by sampling and the
evaluated policy changes. Consequently, \method can use an unfinished
prompt group as a noisy cross-round tail signal without requiring the
regenerated trajectory to reproduce its previous response length.

\section{Controller Implementation and Parameters}
\label{app:controller-calibration}

\method uses a measured inner loop for the tail-group quota $q$ and a
model-based outer loop for the tail-pool replica count $m$.  The inner loop
uses equal bulk and tail completion times as its continuous target at fixed
$m$.  The outer loop runs only after inner stability is certified by the
time-gap tolerance or an adjacent/boundary integer certificate.  It compares
the current split with its adjacent splits, scoring every split after
rebalancing $q$.

\subsection{Measurements and Inner-Loop Stability}

\paragraph{Measurements.}
In round $t$, $T_t^{\mathrm{bulk}}$ is the time until the first $N-q$ of
$N$ fresh groups complete, and $T_t^{\mathrm{tail}}$ is the completion time
of the $q$ tail groups. Only complete, non-transitional rounds at an
unchanged quota contribute to rate estimates and stability checks;
excluded rounds still count toward elapsed execution time. The normalized
time gap is
\begin{equation}
g_t=\frac{T_t^{\mathrm{bulk}}-T_t^{\mathrm{tail}}}
          {\max\{T_t^{\mathrm{bulk}},T_t^{\mathrm{tail}}\}}.
\label{eq:controller-gap-rates}
\end{equation}

\paragraph{Stability.}
The controller uses the EMA rates in Algorithm~\ref{alg:tailsieve-controller}
and treats the inner loop as stable when the absolute mean gap of the
latest $K$ eligible observations at the same quota is within $\epsilon_q$.
For integer quotas, it may also stop at an adjacent endpoint selected by
comparing mean costs when gap signs differ, or at a feasible boundary with
a persistent outward-pointing gap.

\paragraph{Fresh observations.}
After changing $q$ or $m$, the controller restarts the EMA and collects
fresh observations at the new configuration before assessing stability.
Transitional or incomplete rounds also restart this measurement window.
A new workload identity or a substantial change in workload scale starts a
new control epoch with fresh observations.

The 35B, G64 configuration uses the feasible ranges
$\mathcal Q=\{1,\ldots,32\}$ and $\mathcal M=\{1,2,3\}$.

\begin{table}[htbp]
\centering
\small
\caption{Key controller parameters for the 35B, G64 configuration.}
\label{tab:controller-settings}
\begin{tabularx}{0.85\linewidth}{@{}Xl@{}}
\toprule
Parameter & Setting \\
\midrule
Observation window $K$ & $2$ \\
EMA decay $\beta$ & $0.7$ \\
Work-profile window $W$ & $3$ complete batches \\
Balance tolerance $\epsilon_q$ & $0.05$ \\
Replica gain threshold $\epsilon_m$ & $0.01$ \\
\bottomrule
\end{tabularx}
\end{table}

\subsection{Predicting Completion Times}

For each template group $i$, the work estimate $w_i$ is the median total
completion-token count of its accepted observations in the latest $W$
complete batches. The predictor combines these estimates with a fixed,
linearly interpolated per-replica aggregate throughput curve $G(c)$ (tokens/s)
and cyclic group assignment. For a replica with sorted group workloads
$w_{(1)}\le\cdots\le w_{(h)}$ and $w_{(0)}=0$, the predicted completion times are
\begin{equation}
F_{(j)}=\sum_{u=1}^{j}
\frac{(w_{(u)}-w_{(u-1)})(h-u+1)}{G(h-u+1)}.
\label{eq:controller-finish-recurrence}
\end{equation}

For a candidate split $m$, the predictor distributes all $N$ fresh groups
over $R-m$ bulk replicas; the $(N-q)$th predicted completion gives
$\widehat T^{\mathrm{bulk}}$. It then reassigns the $q$ groups predicted to
finish last to $m$ tail replicas and recomputes their completion times;
the latest gives $\widehat T^{\mathrm{tail}}$. Enumerating $q\in\mathcal Q$
yields the rebalanced score in Eq.~\eqref{eq:outer-model-objective}.
The request-profiled throughput curve is evaluated at group concurrency,
so these predictions serve as a ranking surrogate rather than calibrated
wall-clock estimates.

\subsection{Outer Updates and Stopping}

Once fresh observations certify inner stability and a current,
workload-compatible model snapshot is available, the controller selects
the feasible neighbor $m^+\in\{m-1,m+1\}\cap\mathcal M$ with the lowest
rebalanced score in Eq.~\eqref{eq:outer-model-objective}. It moves one
replica only if
\begin{equation}
\frac{\widehat\Phi(m)-\widehat\Phi(m^+)}{\widehat\Phi(m)}>
\epsilon_m.
\label{eq:controller-outer-gain}
\end{equation}
The new quota starts at the predicted minimizer for $m^+$ and is then
rebalanced by the measured inner loop.

The current control episode ends when the inner loop is stable and no
neighbor exceeds the gain threshold under the same valid snapshot;
without one, the controller holds $m$ and waits. A changed workload or
refreshed snapshot can restart adaptation. The fluid-limit analysis in
Appendix~\ref{app:outer-local-search} supports adjacent search under its
assumptions; runtime stopping certifies only observed inner stability and
local stationarity of the predictor.

\begin{figure}[!htbp]
  \centering
  \includegraphics[width=0.55\linewidth]{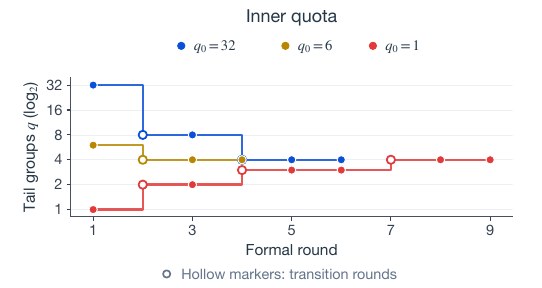}
  \caption{Inner-quota trajectories for Qwen3.5-35B-A3B with four TP2 replicas
  and $q_0=32,6,1$. Hollow markers denote transition rounds.}
  \label{fig:controller-trajectories}
\end{figure}

\paragraph{Controller convergence.}
Figure~\ref{fig:controller-trajectories} shows inner-loop convergence through
round-by-round quota updates. Figure~\ref{fig:spec-convergence}(b) also shows
outer-loop decision convergence to the $2{:}2$ bulk--tail split, with each
point reporting rollout time at the corresponding settled configuration.

\section{Optimality Conditions for Hierarchical Allocation}
\label{app:balancing-proof}

\subsection{Inner-Loop Completion-Time Balance}
\label{app:inner-balance}

We first show the balancing property under a continuous relaxation of the routing problem, where an infinitesimal amount of workload can be shifted between replicas.

Consider two active replicas \(a\) and \(b\), with completion times \(T_a\) and \(T_b\).
Suppose, for contradiction, that an optimal solution satisfies
\[
T_a > T_b .
\]
Let \(\delta = T_a - T_b > 0\).
Under the continuous relaxation, we can shift an infinitesimal amount \(\epsilon\) of workload from the slower replica \(a\) to the faster replica \(b\).
Let \(c_a>0\) denote the marginal decrease in \(T_a\), and let \(c_b>0\) denote the marginal increase in \(T_b\).
For sufficiently small \(\epsilon\), the updated completion times satisfy
\[
T_a' = T_a - c_a\epsilon + o(\epsilon),
\]
and
\[
T_b' = T_b + c_b\epsilon + o(\epsilon).
\]
Since \(T_b<T_a\), we can choose \(\epsilon\) small enough such that
\[
T_b' < T_a .
\]
At the same time,
\[
T_a' < T_a .
\]
Therefore,
\[
\max\{T_a',T_b'\} < T_a = \max\{T_a,T_b\},
\]
which contradicts the optimality of the original solution.
Thus, in the continuous relaxation, an optimal min-max routing solution equalizes the completion times of all active replicas.

In the discrete routing problem, requests are indivisible, so exact equality of replica completion times is not always guaranteed.
Instead, the optimal solution is balanced up to the granularity of movable requests.
Let \(S_r\) denote the set of requests assigned to replica \(r\), and let
\[
M = \max_r T_r(S_r)
\]
be the makespan of a discrete assignment.
For any bottleneck replica \(a\) with \(T_a(S_a)=M\), if there exists a request \(q\in S_a\) and another replica \(b\) such that moving \(q\) from \(a\) to \(b\) yields
\[
T_a(S_a\setminus \{q\}) < M
\]
and
\[
T_b(S_b\cup \{q\}) < M,
\]
while all other replicas remain below \(M\), then the new assignment has a strictly smaller makespan.
This contradicts the optimality of the original assignment.
Therefore, a discrete optimal assignment may not make all replicas finish exactly at the same time, but it admits no workload reassignment that can further reduce the maximum completion time.

\paragraph{Discrete update.}
The current controller projects and clips the EMA-based balancing target,
selects its better floor or ceiling under the min--max rate surrogate,
and restarts its measurement window after a change. It assesses stability
using fresh observations, with finite-integer and boundary safeguards
(Appendix~\ref{app:controller-calibration}).
The balancing condition also explains the responses observed in the oracle routing result.
The oracle does not balance the number of requests across replicas; instead, it assigns a small amount of additional workload to the faster tail replica whenever doing so reduces the bottleneck completion time without making the tail replica the new bottleneck.

\subsection{Outer-Loop Marginal-Capacity Balance}
\label{app:outer-balance}

\paragraph{Continuous condition.}
Let \(p\) be the fraction of routing units assigned to the tail pool
and \(\alpha\) the fraction of replicas serving it. Define
\[
A(p,\alpha)=T^{\mathrm{tail}}(p,\alpha),
\qquad
B(p,\alpha)=T^{\mathrm{bulk}}(p,\alpha).
\]
Locally, moving more traffic to the tail pool makes it slower and the
bulk pool faster, whereas moving more replicas to the tail pool has the
opposite effect:
\[
A_p>0,\qquad B_p<0,\qquad
A_\alpha<0,\qquad B_\alpha>0.
\]

For a fixed \(\alpha\), the inner-loop equilibrium \(p^*(\alpha)\)
satisfies
\[
A\!\left(p^*(\alpha),\alpha\right)
-B\!\left(p^*(\alpha),\alpha\right)=0.
\]
Implicit differentiation gives
\[
\frac{d p^*}{d\alpha}
=-
\frac{A_\alpha-B_\alpha}{A_p-B_p}.
\]
The balanced completion time seen by the outer loop is
\[
\Phi(\alpha)
=A\!\left(p^*(\alpha),\alpha\right)
=B\!\left(p^*(\alpha),\alpha\right).
\]
Therefore,
\[
\frac{d\Phi}{d\alpha}
=A_\alpha+A_p\frac{d p^*}{d\alpha}
=
\frac{A_pB_\alpha-A_\alpha B_p}{A_p-B_p}.
\]
At an interior stationary point, \(d\Phi/d\alpha=0\), and hence
\[
A_pB_\alpha=A_\alpha B_p,
\qquad\text{or equivalently}\qquad
\frac{-A_\alpha}{A_p}
=
\frac{B_\alpha}{-B_p}.
\]
The left ratio is the additional tail traffic that can be absorbed per
incremental increase in tail capacity without changing the tail
completion time. The right ratio is the traffic that must leave the
bulk pool under the corresponding capacity loss. Their equality is the
marginal-capacity balance condition.

To express the condition as a replica fraction, define the per-replica
concurrencies
\[
b_{\mathrm{tail}}=\frac{Np}{R\alpha},
\qquad
b_{\mathrm{bulk}}=\frac{N(1-p)}{R(1-\alpha)},
\]
the local completion-time elasticities
\[
\gamma_{\mathrm{tail}}
=\frac{\partial\ln A}{\partial\ln b_{\mathrm{tail}}},
\qquad
\gamma_{\mathrm{bulk}}
=\frac{\partial\ln B}{\partial\ln b_{\mathrm{bulk}}},
\]
and the marginal cutoff pressures
\[
\lambda_{\mathrm{tail}}
=\frac{\partial\ln A}{\partial p},
\qquad
\lambda_{\mathrm{bulk}}
=-\frac{\partial\ln B}{\partial p}.
\]
At the inner-loop equilibrium, where \(A=B\), these definitions imply
\[
A_p=A\lambda_{\mathrm{tail}},
\qquad
B_p=-B\lambda_{\mathrm{bulk}},
\]
and
\[
A_\alpha=-\frac{A\gamma_{\mathrm{tail}}}{\alpha},
\qquad
B_\alpha=\frac{B\gamma_{\mathrm{bulk}}}{1-\alpha}.
\]
Substitution into the marginal-capacity condition yields
\[
\alpha\lambda_{\mathrm{tail}}\gamma_{\mathrm{bulk}}
=(1-\alpha)\gamma_{\mathrm{tail}}\lambda_{\mathrm{bulk}},
\]
and therefore
\[
\alpha^*
=
\frac{\gamma_{\mathrm{tail}}\lambda_{\mathrm{bulk}}}
{\gamma_{\mathrm{tail}}\lambda_{\mathrm{bulk}}
 +\gamma_{\mathrm{bulk}}\lambda_{\mathrm{tail}}}.
\]

For a homogeneous workload, the two pools have the same local
elasticity and differ only through their per-replica routing-unit counts.
In that limit,
\(\lambda_{\mathrm{tail}}\simeq\gamma/p\) and
\(\lambda_{\mathrm{bulk}}\simeq\gamma/(1-p)\), so the expression
reduces to \(\alpha^*=p\). The two pools then have equal per-replica
loads and are operationally equivalent to uniform routing.

\subsection{Single-Optimum Structure in the Fluid Limit}
\label{app:outer-local-search}

The outer loop changes the replica split after the inner loop has rebalanced
the isolation quota. We analyze the high-concurrency fluid limit and show
that its rebalanced outer objective has a single optimum under the conditions
below, providing structural support for neighboring-split search.

\subsubsection{Fluid Model and Assumptions}
We analyze an ideal oracle partition, where the longest requests are
assigned to the tail pool before execution and bulk executes only the
remaining requests.
Let $p=q/N$ be the longest-request fraction assigned to the tail pool,
$x=m/R$ its replica fraction, and $\rho=N/R$ the average number of requests
per replica. For a response length $L>0$ almost surely, write
$S(z)=\Pr(L>z)$. Let $G(b)>0$ be the differentiable total decoding
throughput of one GPU at concurrency $b>0$, and let $\tau(b)=b/G(b)$
be the corresponding per-token time. Set $\tau(0)=0$, the continuous
extension implied by the elasticity bounds below. Under ideal within-pool
load balancing, the two completion times are
\begin{align}
T_{\mathrm{tail},\rho}(p,x)
&=\int_0^\infty
\tau\!\left(\rho\frac{\min\{p,S(z)\}}{x}\right)dz,\label{eq:outer-tail-fluid}\\
T_{\mathrm{bulk},\rho}(p,x)
&=\int_0^\infty
\tau\!\left(\rho\frac{(S(z)-p)_+}{1-x}\right)dz.\label{eq:outer-bulk-fluid}
\end{align}
For fixed $x$, the tail time increases with $p$, while the bulk time
decreases. Their unique crossing $p_\rho(x)$ minimizes the makespan for that
split. The outer loop therefore sees the rebalanced envelope
\[
\Phi_\rho(x)
=T_{\mathrm{tail},\rho}(p_\rho(x),x)
=T_{\mathrm{bulk},\rho}(p_\rho(x),x).
\]

Define the throughput elasticity
\[
\varepsilon_G(b)=\frac{bG'(b)}{G(b)}.
\]
For every $b>0$, we assume a differentiable, non-increasing elasticity
profile bounded away from zero and one:
\begin{equation}
\varepsilon_G'(b)\le0,
\qquad
0<\underline\alpha
\le\varepsilon_G(b)
\le\overline\alpha<1.
\label{eq:outer-elasticity-condition}
\end{equation}
Hence $\varepsilon_G(b)\to\alpha_\infty\in(0,1)$. Let
$\gamma=1-\alpha_\infty$ and assume
\begin{equation}
\int_0^\infty S(z)^{1-\overline\alpha}\,dz<\infty.
\label{eq:outer-tail-integrability}
\end{equation}
The bounds apply also at small positive concurrency: at any finite $\rho$,
the completion-time integrals can sample arbitrarily small $\rho S(z)$.

\subsubsection{Main Theorem and Proof}

\begin{theorem}[Unique limiting rebalanced outer optimum]
\label{thm:outer-unique-limit}
Suppose that the GPUs are homogeneous, the tail pool receives the longest
$p$ fraction of requests, and the inner quota is rebalanced at every replica
split. Under Eqs.~\eqref{eq:outer-elasticity-condition}--\eqref{eq:outer-tail-integrability},
as $\rho\to\infty$ the normalized
outer objective $\Phi_\rho(x)/\tau(\rho)$ converges uniformly on every compact
subinterval of $(0,1)$ to a limit $\Phi_\infty(x)$. For any non-degenerate
Lognormal, Weibull, Gamma, exponential, Uniform, Pareto, or Lomax
response-length distribution satisfying the conditions in
Table~\ref{tab:outer-uniqueness-distributions}, $\Phi_\infty$ strictly
decreases before one interior point $x^*$ and strictly increases after it.
Thus $x^*$ is the unique global minimizer.
\end{theorem}

\begin{proof}
\textit{Uniform fluid limit.}
Write $\beta_-=1-\overline\alpha$ and
$\beta_+=1-\underline\alpha$, so that
$0<\beta_-\le\gamma\le\beta_+<1$. Since
$d\log\tau(b)/d\log b=1-\varepsilon_G(b)$, integration gives
\[
\log\frac{\tau(\rho u)}{\tau(\rho)}
=\int_1^u[1-\varepsilon_G(\rho v)]\,\frac{dv}{v}.
\]
Consequently, for $r_\rho(u)=\tau(\rho u)/\tau(\rho)$,
\[
\begin{cases}
u^{\beta_+}\le r_\rho(u)\le u^{\beta_-},&0<u\le1,\\
u^{\beta_-}\le r_\rho(u)\le u^{\beta_+},&u\ge1.
\end{cases}
\]
Taking $\rho=1$ also proves the continuous extension $\tau(0)=0$.
As $\rho\to\infty$, the logarithmic integral yields
$r_\rho(u)\to u^\gamma$, uniformly on every interval $[h,M]$ with
$h>0$. This convergence is uniform on $[0,M]$ as well: for $0\le u\le h\le1$,
both $r_\rho(u)$ and $u^\gamma$ are bounded by $h^{\beta_-}$.

Fix $K=[\delta,1-\delta]\subset(0,1)$. For all $p\in[0,1]$ and $x\in K$,
each of $u=\min\{p,S(z)\}/x$ and $u=(S(z)-p)_+/(1-x)$ satisfies
$0\le u\le S(z)/\delta$. Thus both the normalized integrand and its limit
are bounded by the same envelope:
\[
r_\rho(u),\ u^\gamma
\ \le\ \max\{u^{\beta_-},u^{\beta_+}\}
\ \le\ \delta^{-\beta_+}S(z)^{\beta_-}.
\]
Equation~\eqref{eq:outer-tail-integrability} makes this envelope integrable.
Applying dominated convergence to the supremum of the integrand error
over $(p,x)\in[0,1]\times K$ gives uniform convergence of both integrals.
This explains the exponent $1-\overline\alpha$ in the assumption: for
$u<1$, the smaller exponent gives the upper bound.

Define
\[
a(p)=\int_0^\infty\min\{p,S(z)\}^\gamma dz,
\qquad
b(p)=\int_0^\infty\bigl[(S(z)-p)_+\bigr]^\gamma dz.
\]
The two limiting times are $a(p)/x^\gamma$ and $b(p)/(1-x)^\gamma$.
Because $L>0$ almost surely, for every $p<1$ the set $\{z:S(z)>p\}$
has positive measure. The elasticity bounds make $\tau$ strictly increasing,
so both the finite-$\rho$ and limiting balance functions are continuous
and strictly increasing in $p$, with opposite signs at $p=0,1$.
Let $p_\infty(x)$ be the unique root of
$F(p,x)=a(p)/x^\gamma-b(p)/(1-x)^\gamma$.
Joint continuity and strict monotonicity make $p_\infty$ continuous.
For any $\epsilon>0$, $|F|$ has a positive minimum on the compact set
$\{(p,x)\in[0,1]\times K:|p-p_\infty(x)|\ge\epsilon\}$ whenever it is nonempty.
The uniform convergence of the balance functions therefore implies
$p_\rho\to p_\infty$ uniformly on $K$. Substituting these roots into
the uniformly convergent times proves the stated uniform convergence
of $\Phi_\rho/\tau(\rho)$.

\textit{Parameterization and derivative.}
The limiting inner balance has the parameterization
\begin{equation}
x(p)=
\frac{a(p)^{1/\gamma}}
{a(p)^{1/\gamma}+b(p)^{1/\gamma}},
\qquad
\Phi_\infty(x(p))=D(p)^\gamma,
\label{eq:outer-balanced-envelope}
\end{equation}
where
\[
D(p)=a(p)^{1/\gamma}+b(p)^{1/\gamma}.
\]
Because $a$ strictly increases and $b$ strictly decreases, $x(p)$ is a
continuous strictly increasing bijection from $[0,1]$ to $[0,1]$.
It therefore preserves the order and uniqueness of minimizers.

Let $p=S(t)$ and $k=(1-\gamma)/\gamma$. Differentiating $D$ gives
\begin{equation}
D'(p)=t^{1/\gamma}
\left(
\left[1+\frac{e_\gamma(t)}{t}\right]^k
-K_\gamma(t)
\right),
\label{eq:outer-d-prime}
\end{equation}
where
\[
e_\gamma(t)=
\int_t^\infty\left[\frac{S(z)}{S(t)}\right]^\gamma dz.
\]
Let $U\sim\operatorname{Uniform}(0,1)$ and
$Z_t(u)=S(tu)-S(t)$. The normalized bulk-shape term is
\[
K_\gamma(t)=
\left(\mathbb E Z_t(U)^\gamma\right)^{(1-\gamma)/\gamma}
\mathbb E Z_t(U)^{\gamma-1}.
\]
For the listed families, the density $f$ is positive and continuous at
every interior threshold $t$. Hence $S(z)-p\sim f(t)(t-z)$ as $z\uparrow t$,
so the derivative integral below is finite since $\gamma-1>-1$.
In particular,
\[
a(p)=p^\gamma[t+e_\gamma(t)],\qquad
a'(p)=\gamma t p^{\gamma-1},\qquad
b'(p)=-\gamma\int_0^t[S(z)-p]^{\gamma-1}dz,
\]
which verifies Eq.~\eqref{eq:outer-d-prime}, including its positive prefactor.
Lemma~\ref{lem:outer-shape-monotonicity} below shows that
$e_\gamma(t)/t$ is non-increasing and $K_\gamma(t)$ is non-decreasing.
Its family-specific verification gives strict decrease of the first term
for Lognormal, Weibull, Gamma, exponential, Lomax, and Uniform laws,
and strict increase of the second term for Pareto laws.
Consequently, the bracket in Eq.~\eqref{eq:outer-d-prime} strictly decreases
with $t$ and, because $p=S(t)$ decreases with $t$, $D'(p)$ changes sign at
most once, from negative to positive.

\textit{Existence of an interior minimizer.}
Let $M=(\int_0^\infty S(z)^\gamma dz)^{1/\gamma}<\infty$;
finiteness follows from $\gamma\ge\beta_-$ and $0\le S\le1$.
For $p\in(0,1)$, set $f_p=\min\{p,S\}$, $g_p=(S-p)_+$,
$A=a(p)^{1/\gamma}>0$, and $B=b(p)^{1/\gamma}>0$.
Then $f_p+g_p=S$. Apply concavity of $v^\gamma$ to the normalized
functions with $\lambda=A/(A+B)$:
\[
\begin{aligned}
\frac{\int_0^\infty S(z)^\gamma dz}{(A+B)^\gamma}
&=\int_0^\infty
\left(\lambda\frac{f_p(z)}A+(1-\lambda)\frac{g_p(z)}B\right)^\gamma dz\\
&\ge\lambda\int_0^\infty\left(\frac{f_p(z)}A\right)^\gamma dz
+(1-\lambda)\int_0^\infty\left(\frac{g_p(z)}B\right)^\gamma dz=1.
\end{aligned}
\]
Strict concavity gives equality only if $f_p/A=g_p/B$ almost everywhere.
For each listed distribution and every interior $p$, there is a
positive-measure interval beyond $t$ on which $0<S(z)<p$; there
$f_p>0$ and $g_p=0$. Since $g_p$ is positive on another interval,
the normalized functions cannot agree almost everywhere. Therefore
\[
D(p)=A+B<M=D(0)=D(1),\qquad 0<p<1.
\]
The endpoint values follow directly from the definitions of $a,b$.
Moreover, $f_p^\gamma,g_p^\gamma\le S^\gamma$, so dominated convergence
makes $a,b$, and hence $D$, continuous on $[0,1]$.
Thus $D$ attains a global minimum, and the strict inequality excludes
both endpoints. Its derivative must vanish at an interior minimizer.
The strict monotonicity of the bracket in Eq.~\eqref{eq:outer-d-prime}
makes this zero unique, with $D'<0$ before it and $D'>0$ after it.
Equation~\eqref{eq:outer-balanced-envelope} transfers these conclusions
from $p$ to $x$. No convexity of $D$ is required.
\end{proof}

\subsubsection{Supporting Shape Lemma}

\begin{lemma}[Monotonicity of the normalized pool shapes]
\label{lem:outer-shape-monotonicity}
Let $0<\gamma<1$ and let $L$ have a density $f$ supported on an interval
$(\ell,r)\subseteq(0,\infty)$, allowing $\ell=0$ or $r=\infty$.
Suppose $f$ is positive and continuous in the interior,
$\int_0^\infty S(z)^\gamma dz<\infty$, and the density
$g(y)=e^y f(e^y)$ of $\log L$, extended by zero outside its support,
is log-concave. For $\ell<t<r$, $e_\gamma(t)/t$ is non-increasing and
$K_\gamma(t)$ is finite and non-decreasing. The former is strictly
decreasing if $\log g$ is strictly concave on its support. For every family
in Table~\ref{tab:outer-uniqueness-distributions}, at least one of these two
monotonicities is strict.
\end{lemma}

\begin{proof}
\emph{Regularity and the bulk-shape term.}
At an interior $t$, $Z_t(u)\sim t f(t)(1-u)$ as $u\uparrow1$, while
$Z_t(u)\to1-S(t)>0$ as $u\downarrow0$. Thus both moments defining
$K_\gamma$ are finite, including the negative moment because
$\gamma-1>-1$.
Fix $\ell<t_1<t_2<r$, write $Z_i=Z_{t_i}$, and set
$q_i(v)=t_i f(t_i v)=g(\log t_i+\log v)/v$.
Log-concavity gives the ratio ordering, expressed to include zeros below
the lower support endpoint as
\[
q_2(v_1)q_1(v_2)\ge q_1(v_1)q_2(v_2),
\qquad 0<v_1<v_2<1.
\]
Indeed, on the common positive support the ratio $q_2/q_1$ is
non-increasing, since increments of a concave function decrease with its
argument; below that support the inequality follows from the ordered
lower support endpoints $\ell/t_2\le\ell/t_1$.
Since $Z_i(u)=\int_u^1q_i(v)dv$, integration of this inequality gives
\[
\frac{d}{du}\frac{Z_2(u)}{Z_1(u)}
=\frac{q_1(u)Z_2(u)-q_2(u)Z_1(u)}{Z_1(u)^2}\le0
\quad\text{a.e.}
\]
Let $h(u)=\log[Z_2(u)/Z_1(u)]$ and interpolate with
$Z_s(u)=Z_1(u)^{1-s}Z_2(u)^s$, $0\le s\le1$.
Both $Z_s$ and $h$ are non-increasing in $u$. The ratio $Z_2/Z_1$ has
finite positive limits at both endpoints, so $h$ is bounded and the
following differentiation is justified by the moment bounds above.
Write $\mathbb E_{a,s}$ for expectation under the probability density
proportional to $Z_s(u)^a$ on $(0,1)$, and set $k=(1-\gamma)/\gamma$.
Then
\begin{align*}
\frac{d}{ds}\log\!\left[
 (\mathbb E Z_s^\gamma)^k\mathbb E Z_s^{\gamma-1}\right]
&=(1-\gamma)(\mathbb E_{\gamma,s}h-
                    \mathbb E_{\gamma-1,s}h)\\
&=(1-\gamma)
\frac{\operatorname{Cov}_{\gamma-1,s}(Z_s,h)}
     {\mathbb E_{\gamma-1,s}Z_s}\ge0.
\end{align*}
The covariance is nonnegative because, for independent $U,V$ under the
same tilted measure, twice that covariance equals
$\mathbb E[(Z_s(U)-Z_s(V))(h(U)-h(V))]\ge0$.
Integration over $s$ proves $K_\gamma(t_2)\ge K_\gamma(t_1)$.

\emph{The residual-tail term.}
Let $H(y)=S(e^y)$ be the survival function of $\log L$. For interior $y$,
\[
\frac{H(y)}{g(y)}=\int_0^\infty\frac{g(y+v)}{g(y)}dv
\]
is non-increasing by log-concavity of $g$. Hence the hazard $g/H$ is
non-decreasing and $\log H$ is concave. It follows that
$H(y+d)/H(y)$ is non-increasing in $y$ for every $d>0$, and therefore
\[
\frac{e_\gamma(t)}{t}
=\int_1^\infty\left[\frac{S(tv)}{S(t)}\right]^\gamma dv
\]
is non-increasing. If $\log g$ is strictly concave, the hazard is strictly
increasing: for any two interior $y$ values, the integrand in $H/g$
strictly decreases on a positive interval of $v>0$. The survival ratios
then strictly decrease wherever both numerator survival probabilities
are positive, which gives strict decrease of the displayed integral.

\emph{Verification and strictness for the listed families.}
Up to additive constants, $\log g(y)$ equals
\[
\begin{array}{ll}
\text{Lognormal:}&-(y-\mu)^2/(2\sigma^2),\\
\text{Weibull:}&a y-e^{a y}/\lambda^a,\\
\text{Gamma:}&a y-e^y/\theta,\\
\text{Lomax:}&y-(\kappa+1)\log(1+e^y/\lambda).
\end{array}
\]
These functions are strictly concave for the stated positive parameters;
exponential is the Gamma case $a=1$. Thus their residual-tail term is
strictly decreasing. For Uniform $[\ell,r]$, $\log g(y)=y+\mathrm{const}$
on its support and
\[
\frac{e_\gamma(t)}{t}=\frac{r/t-1}{\gamma+1},
\qquad \ell<t<r,
\]
which is strictly decreasing, including $\ell=0$.

For Pareto with lower endpoint $\ell>0$ and exponent $\kappa$,
$\log g(y)=\mathrm{const}-\kappa y$ on $y>\log\ell$ is concave.
The integrability condition gives $\kappa\gamma>1$, and
$e_\gamma(t)/t=1/(\kappa\gamma-1)$ is constant. To prove the needed
strictness of $K_\gamma$, put $p=(\ell/t)^\kappa$ and $C=p^{-1}-1$.
The exact shape, including the lower-support plateau, is
\[
Z_t(u)=pW_C(u),\qquad
W_C(u)=\min\{C,u^{-\kappa}-1\}.
\]
The factor $p$ cancels in $K_\gamma$, but $C$ strictly increases with $t$.
Set $A(C)=\mathbb E W_C^\gamma$, $B(C)=\mathbb E W_C^{\gamma-1}$, and
$\eta(C)=\Pr(U^{-\kappa}-1>C)=(1+C)^{-1/\kappa}>0$.
Differentiating the truncated moments yields
$A'=\gamma C^{\gamma-1}\eta$ and
$B'=(\gamma-1)C^{\gamma-2}\eta$, so
\[
\frac{d}{dC}\log[A(C)^kB(C)]
=(1-\gamma)\eta(C)C^{\gamma-2}
\frac{CB(C)-A(C)}{A(C)B(C)}>0.
\]
Here $CB-A=\mathbb E[W_C^{\gamma-1}(C-W_C)]>0$, since
$0<W_C<C$ on a set of positive measure. Thus $K_\gamma$ is strictly
increasing for Pareto; scaling does not remove its changing
lower-support plateau.
\end{proof}

\subsubsection{Covered Distribution Families}

\begin{table}[!htbp]
\centering
\caption{Response-length families covered by the single-optimum result.}
\label{tab:outer-uniqueness-distributions}
\small
\begin{tabularx}{0.92\linewidth}{@{}lX@{}}
\toprule
Distribution & Condition \\
\midrule
Lognormal & Non-zero variance \\
Weibull, Gamma, exponential & Positive distribution parameters \\
Uniform $[a,b]$ & $0\le a<b$ \\
Pareto & Lower endpoint $\ell>0$ and $\kappa(1-\overline\alpha)>1$ \\
Lomax & Scale $\lambda>0$ and $\kappa(1-\overline\alpha)>1$ \\
\bottomrule
\end{tabularx}
\end{table}

For Pareto with lower endpoint $\ell>0$ and Lomax with scale $\lambda>0$,
respectively, direct integration gives, for any $\beta>0$,
\[
\int_0^\infty S(z)^\beta dz=
\begin{cases}
\ell+\ell/(\kappa\beta-1),&\text{Pareto},\\
\lambda/(\kappa\beta-1),&\text{Lomax},
\end{cases}
\qquad\text{if }\kappa\beta>1,
\]
and the integral is infinite otherwise. Thus the table's condition is
equivalent to Eq.~\eqref{eq:outer-tail-integrability} for these two families.
It also implies $\kappa\gamma>1$, as required by the limiting integrals.
The latter condition alone need not provide an integrable envelope for
the finite-$\rho$ model.

\subsubsection{Implications for Neighboring Replica Search}

\begin{corollary}[Neighboring replica search in the fluid limit]
Sample the limiting objective on the feasible ordered replica splits
$x_j=j/R$. If there is no exact tie, the resulting sequence strictly
decreases before its unique minimizer and strictly increases afterward.
Starting from any split, repeatedly moving to a lower-valued neighboring
split reaches the global minimizer.
\end{corollary}

This limiting result motivates the controller's local search. \method
evaluates the current and feasible adjacent replica splits, rebalances $q$
for each split in the model, and applies at most one-replica move per
decision. The result supports this search only under the stated fluid-limit
assumptions; the runtime stopping rule is not a finite-workload optimum proof.

\paragraph{Relation to runtime execution.}
At runtime, the bulk pool submits $N$ candidate prompt groups, and
$T^{\mathrm{bulk}}$ is measured when the first $N-q$ groups have completed.
This time includes admission waiting and contention from unaccepted
candidates before the cutoff. Together with the measured tail completion
time, it drives the inner quota loop; the outer replica loop uses the passive
predictor in Appendix~\ref{app:controller-calibration}.

\section{Offline Routing Model and Tail Isolation}
\label{app:offline-routing-analysis}

\subsection{Exact Makespan}

For a routing policy \(p\in\{\mathrm{base},\mathrm{tail}\}\), let
\(S_r^p\) denote the set of requests assigned to replica \(r\).
The active batch size at decoding position \(x\) is
\[
b_r^p(x)
=
\sum_{i\in S_r^p}
\mathbf{1}\{L_i\ge x\}.
\]
Under the all-admit decoding model, the completion time of replica \(r\)
is
\[
T_r^p
=
\sum_{x=1}^{L_{r,\max}^p}
\tau\!\left(b_r^p(x)\right),
\qquad
L_{r,\max}^p
=
\max_{i\in S_r^p}L_i.
\]

Equivalently, the same wall-clock time can be distributed evenly among
the requests active at each decoding position:
\[
T_r^p
=
\sum_{i\in S_r^p}
\sum_{x=1}^{L_i}
\frac{
\tau\!\left(b_r^p(x)\right)
}{
b_r^p(x)
}.
\]
To see the equivalence, note that exactly \(b_r^p(x)\) requests are
active at position \(x\). Therefore,
\[
\sum_{i\in S_r^p}
\mathbf{1}\{L_i\ge x\}
\frac{\tau\!\left(b_r^p(x)\right)}{b_r^p(x)}
=
\tau\!\left(b_r^p(x)\right).
\]
Thus, the factor \(1/b_r^p(x)\) prevents the same concurrent
wall-clock interval from being counted once for every active request.

The exact rollout-step makespan is determined by the slowest replica:
\[
T_{\mathrm{step}}^p
=
\max_r T_r^p,
\]
and the exact gain of tail routing is
\[
\Delta T_{\mathrm{step}}(q,m)
=
T_{\mathrm{step}}^{\mathrm{base}}
-
T_{\mathrm{step}}^{\mathrm{tail}}(q,m).
\]

\subsection{Tail Workload and Concurrency Effects}
\label{app:conditional-gain}

Let \(\mathcal{T}_q\) denote the isolated tail requests.
For request \(i\), let \(b_i^p(x)\) denote the active batch size of the
replica serving it under policy \(p\); under tail routing, this batch size
depends on \((q,m)\). With \([z]_+=\max\{z,0\}\), the tail-side proxy
decomposes into two competing effects:
\[
\begin{aligned}
\widetilde{\Delta T}_{\mathrm{tail}}(q,m)
&=
\underbrace{
\sum_{i\in\mathcal{T}_q}
\sum_{x=1}^{L_i}
\left[
\tau\!\left(b_i^{\mathrm{base}}(x)\right)
-
\tau\!\left(b_i^{\mathrm{tail}}(x;q,m)\right)
\right]_+
}_{\text{low-concurrency advantage}}
\\[1ex]
&\quad-
\underbrace{
\sum_{i\in\mathcal{T}_q}
\sum_{x=1}^{L_i}
\left[
\tau\!\left(b_i^{\mathrm{tail}}(x;q,m)\right)
-
\tau\!\left(b_i^{\mathrm{base}}(x)\right)
\right]_+
}_{\text{tail-concentration penalty}}.
\end{aligned}
\]
This identity holds for any profiled latency curve. The concurrency-based
interpretation below applies where \(\tau(b)\) is nondecreasing.

\paragraph{Low-concurrency advantage.}
The first term accumulates the reduction in per-token decoding latency
experienced by isolated tail requests. Placing long requests on lightly
loaded replicas can lower their decoding latency. This advantage typically
arises early, while the baseline replicas still serve many short requests.

\paragraph{Tail-concentration penalty.}
The second term accumulates the additional per-token latency incurred
when isolated tail requests decode more slowly than under the baseline.
As short requests finish, baseline concurrency falls, whereas concentrating
long requests on only \(m\) replicas can sustain higher tail-side
concurrency. The resulting penalty can offset the initial advantage.
The decomposition separates positive and negative contributions; it does
not require a single early-to-late crossover.

The response-length distribution determines both the total isolated work,
\[
W_{\mathrm{tail}}(q)=\sum_{i\in\mathcal{T}_q}L_i,
\]
and how long requests remain active. The concurrency trajectories
\(b_i^p(x)\), together with \(\tau(b)\), determine the latency difference
at each decoding position. Thus, a heavier tail can extend the work that
benefits from isolation, but can also prolong the concentration penalty.
Increasing \(q\) changes the isolated workload, while changing \(m\) at fixed \(q\)
redistributes that workload across tail replicas. Their best settings
therefore depend jointly on tail shape and concurrency-dependent latency,
and cannot in general be inferred from tail strength alone.

Unlike the exact per-request decomposition above, this proxy does not
divide each shared decoding interval by the number of active requests.
It therefore measures aggregate concurrency exposure over the isolated
tail set rather than exact wall-clock time. Moreover, the exact
rollout-step objective takes the maximum completion time across all
replicas, whereas the proxy only describes the isolated tail requests.

We use this proxy to interpret the two effects above. All oracle values and reported
speedups are computed using the exact replica-level makespan or measured
end-to-end execution time.

\begin{figure}[!htbp]
    \centering
    \includegraphics[width=0.9\linewidth]{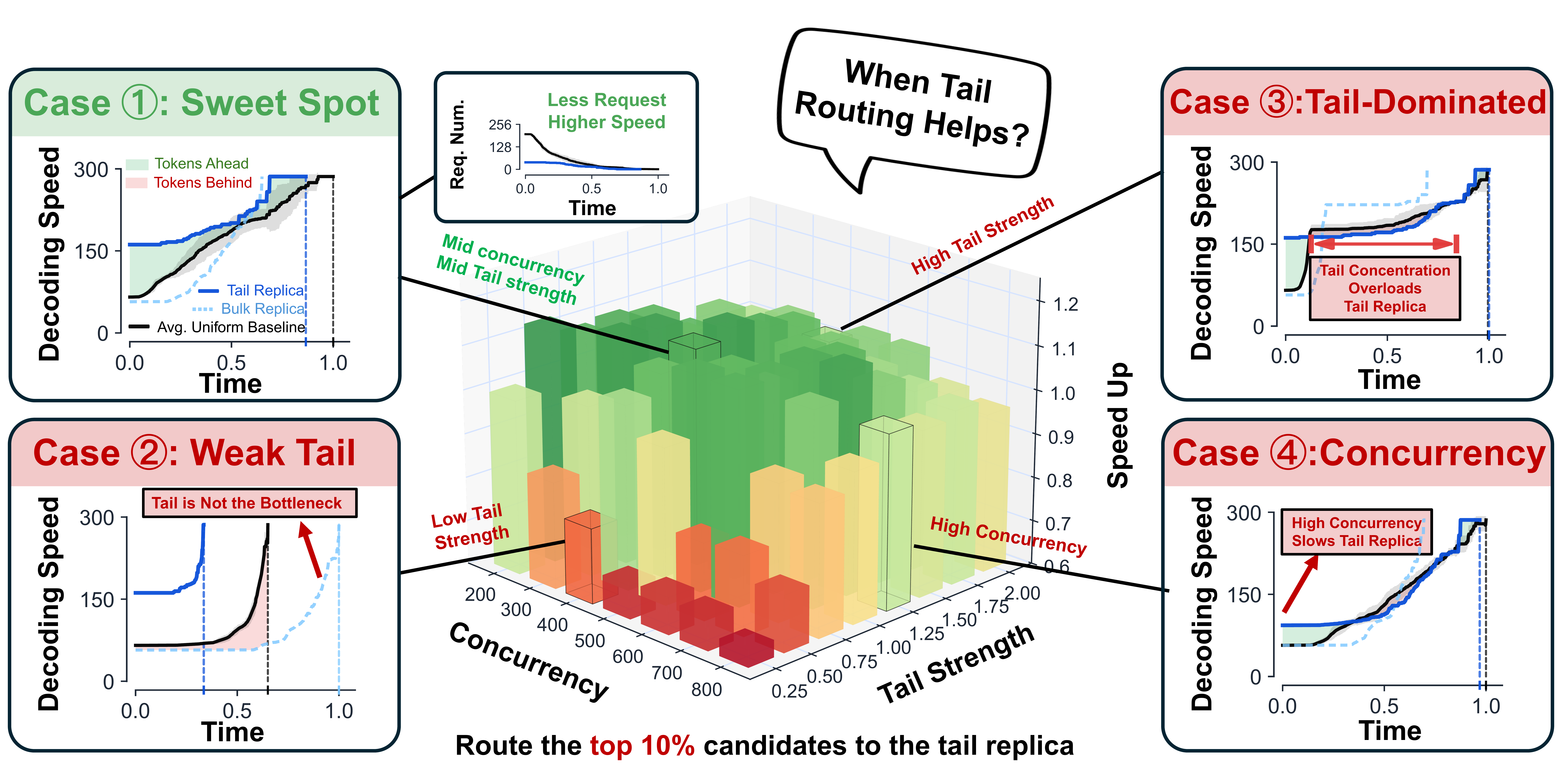}
    \caption{Gain regimes for fixed top-\(10\%\) isolation under a fixed
    replica split.}
    \label{fig:conditional-gain-cases}
\end{figure}

Figure~\ref{fig:conditional-gain-cases} illustrates four representative
regimes. Routing gains weaken when the tail workload is too small, the
tail route becomes overloaded, or isolation loses its low-concurrency
advantage.

\section{Analysis of Routing-Induced Length Bias}
\label{app:length-distribution}

We distinguish natural policy-induced distribution drift from bias
introduced by routing. The marginal response-length distribution may
change as the policy evolves; our question is whether the cross-round
routing pipeline shifts the distribution relative to generation under
the same current policy and prompt stream. We first analyze an idealized
stationary setting with a fixed policy, which isolates the effect of
request composition. Let
\[
\mathcal{X}_t
=
\{X_{t,1},\ldots,X_{t,N}\}
\]
be a fresh pool of \(N\) requests with prompt distribution \(P_X\). We
require consecutive pools to have the same marginal law,
\(\mathcal{X}_t\overset{d}{=}\mathcal{X}_{t'}\), but do not require
them to be independent across steps. Thus, the result also covers, for
example, a stationary ordered or without-replacement prompt stream. For
a fixed policy \(\pi\), let
\(Y(X;\pi,\xi)\) denote the response generated from request \(X\) with
fresh sampling randomness \(\xi\), and let \(L(Y)\) denote its length.
For any measurable length interval \(A\), define the prompt-conditional
probability
\[
p_A(X)
=
\Pr_{\xi}\!\left[L(Y(X;\pi,\xi))\in A\mid X\right]
\]
and the population response-length distribution
\[
P_{\pi}(A)
=
\mathbb{E}_{X\sim P_X}[p_A(X)].
\]

For a stable allocation, let \(n^{\mathrm{tail}}+n^{\mathrm{bulk}}=N\).
We idealize the tail selector as prompt-conditioned: applying the same
selection rule to each pool partitions it into
\[
\mathcal{X}_t
=
\mathcal{T}_{n^{\mathrm{tail}}}(\mathcal{X}_t)
\uplus
\mathcal{F}_{n^{\mathrm{bulk}}}(\mathcal{X}_t),
\]
where the two sets contain \(n^{\mathrm{tail}}\) and
\(n^{\mathrm{bulk}}\) requests, respectively. The selector need not
identify the true longest requests; it only needs to depend on the prompt
associated with each request and remain unchanged across identically
distributed pools. This idealization captures the intuition that
long-tail prompts tend to remain long-tailed, while separating that
prompt-level signal from trajectory-level sampling noise.

Let \(\mathcal{G}_{\pi}(\mathcal{S})\) denote independent generation
under \(\pi\) for every request in \(\mathcal{S}\). For a request set
\(\mathcal{S}\), define its expected number of responses in \(A\) as
\[
C_A(\mathcal{S})
=
\sum_{X\in\mathcal{S}}p_A(X).
\]
For a generated update batch \(\mathcal{B}_t\), define
\[
\widehat P_{\mathcal{B}_t}(A)
=
\frac{1}{N}
\sum_{Y\in\mathcal{B}_t}\mathbf{1}\{L(Y)\in A\}.
\]
Unless stated otherwise, expectations below are joint over the
request pool, fresh generation randomness, and any prompt-conditioned
randomness used by the selector. The equality is therefore a population
statement: a realized finite batch can deviate from it through sampling
variance.

For a fixed-size prompt group, the same result follows by applying this
argument to each response in the group and averaging over the group.
\method always routes and regenerates the complete group together.

\paragraph{Stable allocation.}

When the two allocations remain fixed, the tail route regenerates the
selected requests from the preceding pool, while the bulk route generates
the complementary requests from the current pool. The update batch is
therefore
\[
\mathcal{B}_t
=
\mathcal{G}_{\pi}\!\left(
\mathcal{T}_{n^{\mathrm{tail}}}(\mathcal{X}_{t-1})
\right)
\uplus
\mathcal{G}_{\pi}\!\left(
\mathcal{F}_{n^{\mathrm{bulk}}}(\mathcal{X}_t)
\right).
\]

Since \(\mathcal{X}_{t-1}\) and \(\mathcal{X}_t\) are identically
distributed and use the same prompt-conditioned selector,
\[
\begin{aligned}
\mathbb{E}
\left[
\widehat P_{\mathcal{B}_t}(A)
\right]
&=\frac{1}{N}
\mathbb{E}
\left[
C_A\!\left(\mathcal{T}_{n^{\mathrm{tail}}}(\mathcal{X}_{t-1})\right)
+
C_A\!\left(\mathcal{F}_{n^{\mathrm{bulk}}}(\mathcal{X}_t)\right)
\right]
\\
&=\frac{1}{N}
\mathbb{E}
\left[
C_A\!\left(\mathcal{T}_{n^{\mathrm{tail}}}(\mathcal{X}_t)\right)
+
C_A\!\left(\mathcal{F}_{n^{\mathrm{bulk}}}(\mathcal{X}_t)\right)
\right]
\\
&=\frac{1}{N}
\mathbb{E}
\left[
C_A(\mathcal{X}_t)
\right]
=P_{\pi}(A).
\end{aligned}
\]
Thus, repeated generations of the same prompt or prompt group need not
have identical lengths. Under the stated fixed-policy idealization,
fresh regeneration and cross-round recombination do not introduce
routing-induced length bias: \(\widehat P_{\mathcal{B}_t}(A)\) is an
unbiased estimator of the same-policy population \(P_{\pi}(A)\) for
every measurable interval \(A\). This comparison is against generation
under the same policy; it does not assert that the population
distribution remains unchanged across policy updates.

\paragraph{Increasing the tail allocation.}

Suppose the load balancer increases the tail allocation by \(\Delta>0\):
\[
n^{\mathrm{tail},+}=n^{\mathrm{tail}}+\Delta,
\qquad
n^{\mathrm{bulk},+}=N-n^{\mathrm{tail},+}.
\]
The new allocations are applied when partitioning \(\mathcal{X}_t\).
After one pipeline step, the update batch takes the canonical
request-composition form
\[
\mathcal{B}_{t+1}
=
\mathcal{G}_{\pi}\!\left(
\mathcal{T}_{n^{\mathrm{tail},+}}(\mathcal{X}_t)
\right)
\uplus
\mathcal{G}_{\pi}\!\left(
\mathcal{F}_{n^{\mathrm{bulk},+}}(\mathcal{X}_{t+1})
\right).
\]
Applying the stable-allocation argument gives
\[
\boxed{
\mathbb{E}
\left[
\widehat P_{\mathcal{B}_{t+1}}(A)
\right]
=
P_{\pi}(A)
}.
\]
Provided that no new allocation change is applied during settlement,
the canonical prompt composition is restored one step after increasing
the tail allocation, eliminating the routing-induced deviation in this
idealized setting.

\paragraph{Decreasing the tail allocation.}

Suppose the load balancer decreases the tail allocation by \(\Delta>0\):
\[
n^{\mathrm{tail},-}=n^{\mathrm{tail}}-\Delta,
\qquad
n^{\mathrm{bulk},-}=N-n^{\mathrm{tail},-}.
\]
At the time of this decision, the existing tail-request pool was
constructed using the previous tail allocation \(n^{\mathrm{tail}}\)
and therefore contains \(\Delta\) more requests than the new tail
replica requires. Assuming a nested selector, the old tail set can be
decomposed as
\[
\mathcal{T}_{n^{\mathrm{tail}}}(\mathcal{X}_t)
=
\mathcal{T}_{n^{\mathrm{tail},-}}(\mathcal{X}_t)
\uplus
\mathcal{E}_{\Delta}(\mathcal{X}_t),
\]
where \(\mathcal{E}_{\Delta}\) contains the \(\Delta\) residual requests.

Since \method does not drop requests, these residual requests must first
be drained or reassigned to the bulk replica. Consequently, the first
transition batch after the allocation reduction does not yet have the
canonical new-allocation request composition
\[
\mathcal{T}_{n^{\mathrm{tail},-}}(\mathcal{X})
\uplus
\mathcal{F}_{n^{\mathrm{bulk},-}}(\mathcal{X}').
\]
During this transition step, however, the next fresh pool
\(\mathcal{X}_{t+1}\) is partitioned using the new allocations,
producing
\[
\mathcal{T}_{n^{\mathrm{tail},-}}(\mathcal{X}_{t+1}).
\]

After the residual requests have been drained, the following update
batch is
\[
\mathcal{B}_{t+2}
=
\mathcal{G}_{\pi}\!\left(
\mathcal{T}_{n^{\mathrm{tail},-}}(\mathcal{X}_{t+1})
\right)
\uplus
\mathcal{G}_{\pi}\!\left(
\mathcal{F}_{n^{\mathrm{bulk},-}}(\mathcal{X}_{t+2})
\right).
\]
Applying the same argument gives
\[
\boxed{
\mathbb{E}
\left[
\widehat P_{\mathcal{B}_{t+2}}(A)
\right]
=
P_{\pi}(A)
}.
\]
Provided that no new allocation change is applied during settlement,
the canonical prompt composition is restored two steps after decreasing
the tail allocation, eliminating the routing-induced deviation in this
idealized setting.

\paragraph{Fresh sampling and evolving policies.}

The exact result above uses a stationary policy and a selector that is a
function of the prompt. In practice, the marginal target
\(P_{\pi_t}\) may change from one policy snapshot to the next, and the
partial-rollout selector also depends on a sampled trajectory. The
relevant quantity is therefore the deviation of the routed batch from
the current-policy target \(P_{\pi_t}\), rather than its difference from
the preceding step's distribution. Write
\(Y_t=Y(X;\pi_t,\xi_t)\) for the response generated from request \(X\)
under \(\pi_t\). Let
\(Z_t\in\{0,1\}\) indicate whether request \(X\) is selected as a
tail candidate at step \(t\), and define
\[
s_t(X)=\mathbb{E}[Z_t\mid X],
\qquad
p_{t,A}(X)
=
\Pr\!\left[L(Y_t)\in A\mid X\right].
\]
Two quantities characterize the deviation from the ideal model:
\[
\epsilon_t
=
\mathbb{E}_X\!\left[|s_t(X)-s_{t-1}(X)|\right]
\]
measures cross-policy selector drift, while
\[
\eta_t(A)
=
\mathbb{E}_X\!\left[
\left|
\operatorname{Cov}\!\left(
Z_t,\mathbf{1}\{L(Y_t)\in A\}\mid X
\right)
\right|
\right]
\]
measures residual trajectory dependence after conditioning on the
prompt. For a stable allocation, conditional independence between the
regenerated tail response and the preceding selection gives
\[
\begin{aligned}
\mathbb{E}\!\left[\widehat P_{\mathcal{B}_t}(A)\right]
-P_{\pi_t}(A)
&=
\mathbb{E}_X\!\left[
p_{t,A}(X)\bigl(s_{t-1}(X)-s_t(X)\bigr)
\right]\\
&\quad-
\mathbb{E}_X\!\left[
\operatorname{Cov}\!\left(
Z_t,\mathbf{1}\{L(Y_t)\in A\}\mid X
\right)
\right].
\end{aligned}
\]
The triangle inequality then yields
\[
\left|
\mathbb{E}\!\left[\widehat P_{\mathcal{B}_t}(A)\right]
-P_{\pi_t}(A)
\right|
\leq
\epsilon_t+\eta_t(A).
\]
The trace aggregates in Appendix~\ref{app:prompt-stability} support
using prompt identity as a cross-checkpoint signal, but they do not estimate
$\epsilon_t$ or $\eta_t(A)$. Thus prompt-dominant length variation supports
the selector while the bound keeps its distributional effect explicit.

\paragraph{Summary of distributional conditions.}

The analysis distinguishes three regimes. A stationary policy with
identically distributed prompt pools, a stable allocation, and the same
prompt-conditioned selector introduces no routing-induced bias in
expectation. A stable allocation with policy or selector drift gives the
approximation bound above relative to the current-policy target. An
allocation change creates a finite composition transition, during which
neither equality is claimed. For every measurable length interval
\(A\), the exact fixed-policy special case is
\[
\begin{cases}
\mathbb{E}[\widehat P_{\mathcal{B}_t}(A)]
=
P_{\pi}(A),
&
\text{under stable allocations},\\[4pt]
\mathbb{E}[\widehat P_{\mathcal{B}_{t+1}}(A)]
=
P_{\pi}(A),
&
\text{one step after increasing the tail allocation},\\[4pt]
\mathbb{E}[\widehat P_{\mathcal{B}_{t+2}}(A)]
=
P_{\pi}(A),
&
\text{two steps after decreasing the tail allocation}.
\end{cases}
\]
These equalities compare each routed batch with generation under the same
fixed policy; the response-length distribution may still evolve during
training. Both timelines start with the allocation change applied before
step \(t\), require successful rounds, and assume no further allocation
change through \(t+1\) for an increase or \(t+2\) for a decrease.
Under policy or selector drift, the remaining deviation is governed by
$\epsilon_t+\eta_t(A)$.

\section{Additional Empirical Evaluation}
\label{app:additional-empirical-evaluation}

\subsection{Routing Gains with Matched Speculative Decoding}
\label{app:matched-speculation}
Table~\ref{tab:matched-speculation} separates the combined speedup over
non-speculative Uniform routing from the incremental gain over Uniform
routing with the same speculative decoder (pure MTP or pure DFlash).
All configurations use 64 prompts with eight responses per prompt and
eight GPUs: four TP2 replicas for Qwen3.5-35B-A3B and eight TP1 replicas
for Qwen3.5-4B and Qwen3.5-2B. Routing fixes equal-sized replica pools
and adapts $q$. Both pools use proposal depth three; DFlash's query block
size is four. These comparisons hold proposal depth fixed across routes.

\begin{table}[!htbp]
\centering
\caption{Generation time in seconds and speedup with matched speculative
decoding. All models are Qwen3.5. $T_0$ is Uniform without speculation,
$T_s$ is Uniform with speculation, and $T_r$ is \method with the same
speculation. Routing gain is $T_s/T_r$; combined gain is $T_0/T_r$.}
\label{tab:matched-speculation}
\footnotesize
\begin{tabular*}{\linewidth}{@{\extracolsep{\fill}}llrrrrr@{}}
\toprule
Model & Decoder & $T_0$ & $T_s$ & $T_r$ & Routing gain & Combined gain \\
\midrule
\multicolumn{7}{l}{\emph{DeepScaleR}} \\
35B-A3B & MTP    & 150.27 & 72.44 & 62.87 & 1.152$\times$ & 2.390$\times$ \\
4B      & MTP    &  99.94 & 48.15 & 41.39 & 1.163$\times$ & 2.414$\times$ \\
2B      & MTP    &  62.72 & 38.57 & 29.02 & 1.329$\times$ & 2.161$\times$ \\
35B-A3B & DFlash & 150.27 & 76.27 & 68.91 & 1.107$\times$ & 2.181$\times$ \\
4B      & DFlash &  99.94 & 53.47 & 47.79 & 1.119$\times$ & 2.091$\times$ \\
\midrule
\multicolumn{7}{l}{\emph{KodCode-10K}} \\
35B-A3B & MTP    & 88.16 & 42.01 & 34.09 & 1.232$\times$ & 2.586$\times$ \\
4B      & MTP    & 71.86 & 31.43 & 29.84 & 1.053$\times$ & 2.408$\times$ \\
2B      & MTP    & 56.97 & 30.90 & 24.26 & 1.274$\times$ & 2.349$\times$ \\
35B-A3B & DFlash & 88.16 & 41.41 & 37.07 & 1.117$\times$ & 2.378$\times$ \\
4B      & DFlash & 71.86 & 44.65 & 33.45 & 1.335$\times$ & 2.148$\times$ \\
\bottomrule
\end{tabular*}
\end{table}

With matched speculative decoding, \method achieves up to
1.33$\times$ speedup with MTP and 1.34$\times$ with DFlash over uniform
routing.

\subsection{Downstream Evaluation and Uncertainty}
\label{app:downstream-uncertainty}

GSM8K and MATH-500 use all 1,319 and 500 test questions, respectively,
with greedy decoding (temperature~0, one response per question). Each
AIME year contains 30 questions. For every checkpoint and year, we use
three evaluation seeds, with ten responses per question
and seed, temperature~0.6, top-$p=0.95$, and top-$k$ filtering disabled
(\texttt{top\_k=0}). The three seeds use non-overlapping response-sampling
streams and are fixed across checkpoints.
All evaluations disable thinking, stop at \texttt{<END>}, and cap responses
at 16,384 tokens. An observed pass@10 score marks a question correct when
at least one of its ten responses is correct. Table~\ref{tab:rl-accuracy}
reports the mean of the three AIME scores; Table~\ref{tab:accuracy-seeds}
reports the per-seed results and their sample standard deviation.

\begin{table}[!htbp]
\centering
\caption{AIME pass@10 (percent) by evaluation sampling seed; SD is across three seeds, with each checkpoint fixed.}
\label{tab:accuracy-seeds}
\scriptsize
\setlength{\tabcolsep}{4pt}
\begin{tabular}{lllrrrr}
\toprule
Model & Method & Dataset & Seed 1 & Seed 2 & Seed 3 & Mean $\pm$ SD \\
\midrule
2B & Base & AIME24 & 30.00 & 33.33 & 36.67 & $33.33\pm3.33$ \\
2B & Base & AIME25 & 23.33 & 26.67 & 30.00 & $26.67\pm3.33$ \\
2B & Baseline & AIME24 & 50.00 & 63.33 & 56.67 & $56.67\pm6.67$ \\
2B & Baseline & AIME25 & 36.67 & 36.67 & 36.67 & $36.67\pm0.00$ \\
2B & Routing-only & AIME24 & 53.33 & 60.00 & 60.00 & $57.78\pm3.85$ \\
2B & Routing-only & AIME25 & 36.67 & 43.33 & 43.33 & $41.11\pm3.85$ \\
2B & Routing + MTP & AIME24 & 46.67 & 53.33 & 60.00 & $53.33\pm6.67$ \\
2B & Routing + MTP & AIME25 & 33.33 & 43.33 & 33.33 & $36.67\pm5.77$ \\
\midrule
4B & Base & AIME24 & 90.00 & 93.33 & 86.67 & $90.00\pm3.33$ \\
4B & Base & AIME25 & 83.33 & 83.33 & 80.00 & $82.22\pm1.92$ \\
4B & Baseline & AIME24 & 90.00 & 93.33 & 90.00 & $91.11\pm1.92$ \\
4B & Baseline & AIME25 & 83.33 & 80.00 & 76.67 & $80.00\pm3.33$ \\
4B & Routing-only & AIME24 & 93.33 & 90.00 & 90.00 & $91.11\pm1.92$ \\
4B & Routing-only & AIME25 & 86.67 & 80.00 & 80.00 & $82.22\pm3.85$ \\
4B & Routing + MTP & AIME24 & 90.00 & 90.00 & 90.00 & $90.00\pm0.00$ \\
4B & Routing + MTP & AIME25 & 83.33 & 86.67 & 83.33 & $84.44\pm1.92$ \\
\bottomrule
\end{tabular}
\end{table}

The across-seed standard deviations describe evaluation sampling
variation at fixed checkpoints and questions.

\subsection{Accepted-Group Length Distributions}
\label{app:accepted-group-length-distributions}
\begin{figure}[htbp]
    \centering
    \includegraphics[width=\textwidth]{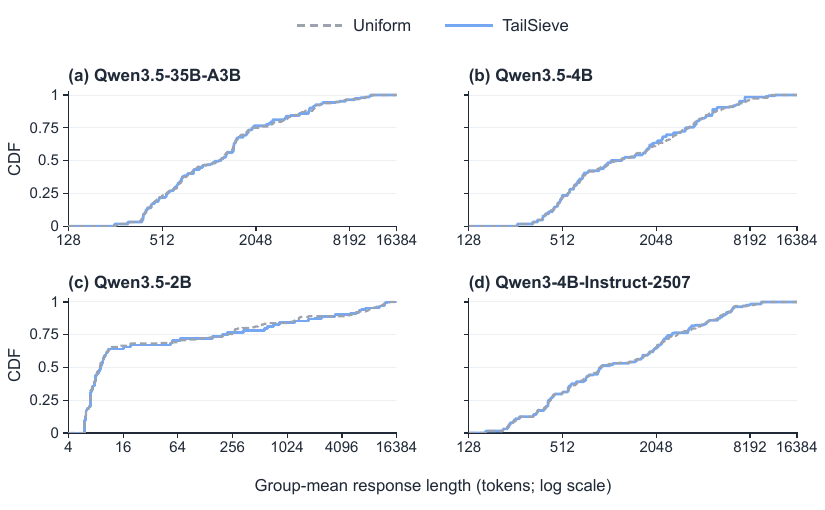}
    \caption{Empirical CDFs of mean response length per accepted prompt
    group on DeepScaleR. Each curve pools three evaluation batches
    of 64 groups, with eight responses per group. Group mean length is the
    accepted completion-token total divided by eight. Both methods use the
    same prompt pool and sampling settings; horizontal axes are logarithmic.}
    \label{fig:deepscaler-group-length-cdf}
\end{figure}

Figure~\ref{fig:deepscaler-group-length-cdf} compares Uniform and \method
on DeepScaleR across four model configurations. The maximum absolute
gap between their empirical CDFs of accepted-group mean response lengths
is below 5.3\% in every configuration.

\section{Detailed Related Work}
\label{app:detailed_related_work}
\label{app:related-work}

\paragraph{Tail latency and straggler mitigation.}
Tail latency is a longstanding concern in distributed systems because a
small number of slow tasks can determine end-to-end completion time
~\citep{dean2013tail}. Classical mitigations use hedged execution or
task cloning to mask stragglers with redundant work
~\citep{ananthanarayanan2013clones}. Long generations in synchronous
LLM RL are instead workload-level stragglers: their useful responses
must eventually be collected rather than raced against identical
copies. \method therefore reshapes request placement and replica
capacity instead of relying on full-rollout duplication.

\paragraph{Long-tail rollout execution in LLM RL.}
Existing RL systems address rollout imbalance by relaxing or
reorganizing synchronization. AReaL decouples rollout generation from
training~\citep{fu2025areal}; Kimi k1.5 reuses trajectory segments
~\citep{kimi2025k15}; and APRIL and CoPRIS overprovision rollouts and
carry unfinished work across steps~\citep{zhouapril2025,qu2025copris}.
RollPacker preserves synchronous training but consolidates predicted
tails into tail-heavy rounds~\citep{gao2026rollpacker}. In contrast,
\method uses cutoff status only as a tail signal, regenerates every
consumed response under the current policy, and preserves mixed
tail--bulk updates while reducing each rollout barrier.

\paragraph{Length-aware routing and adaptive capacity allocation.}
Length-aware serving schedulers estimate completion lengths using proxy
models, predictive distributions, or entropy-guided representations
~\citep{qiu2024ssjf,zheng2026tie,xie2026forelen}. For RL rollouts,
StreamRL learns a length ranker for skew-aware dispatch
~\citep{zhong2025streamrl}, while Seer exploits within-prompt response
similarity through online context learning and divided rollout
~\citep{qin2026seer}. Libra maintains residual-length histograms in a prefix tree of
tool-execution states to guide request migration among tensor-parallel
buckets~\citep{chen2026libra}. Its global planner also adjusts stage
budgets and parallelism configurations. \method instead treats the isolated
workload $q$ and tail-pool replica count $m$ as a joint control problem
within synchronous rollout, using partial-rollout observations,
response-work history, and measured concurrency--throughput behavior
without requiring exact length predictions. It combines this allocation
with route-specific speculation.

\paragraph{Speculative decoding and draft models.}
Speculative decoding accelerates autoregressive generation by using a cheap proposal mechanism and then verifying the proposed tokens with the target model~\citep{leviathan2023fast,chen2023accelerating}.
Early formulations use an independent smaller draft model, while later work improves candidate quality through blockwise prediction, tree verification, and feature-level drafting~\citep{stern2018blockwise,miao2024specinfer,li2024eagle,li2024eagle2,li2025eagle}.
Self-speculative decoding, where the target model reuses its own intermediate layers for drafting and verification, eliminating the need for a separate draft model; this direction was initiated by Draft \& Verify and later extended to on-the-fly and dynamically optimized variants such as SWIFT and KNN-SSD~\citep{zhang2024draft,xia2025swiftontheflyselfspeculativedecoding,song-etal-2026-knn}.
Medusa-style methods attach auxiliary decoding heads to the target model and generate multiple future-token proposals in parallel~\citep{cai2024medusa}.
EAGLE-style methods instead reuse target hidden features and construct a draft tree, offering a strong practical baseline for lossless LLM acceleration~\citep{li2024eagle,li2024eagle2,li2025eagle}.

\paragraph{Dynamic tree construction and adaptive speculation.}
Tree-based speculative decoding increases the chance of accepting multiple tokens by verifying multiple candidate paths in one target forward~\citep{miao2024specinfer}.
Static tree methods are easy to deploy but may waste budget on low-confidence branches.
Adaptive speculation methods adjust the draft length or tree topology based on confidence, acceptance history, or token probabilities~\citep{mamou2024dynamic,zhang2024adaeagle,brown2024dynamic,hu2026echo,shen2026draftlessretrievemore}.
These methods expose a central trade-off: pruning or early stopping can reduce draft-side overhead, but it may also remove valid continuations and lower MAT.
ECHO further studies this issue in high-concurrency settings and formulates elastic budget scheduling across requests~\citep{hu2026echo}.

\paragraph{Retrieval-based speculative decoding.}
Retrieval is an attractive proposal source because repeated local patterns can be reused with little draft-model computation.
Lookahead decoding, PLD, REST, Token Recycling, LogitSpec, SAMD, and Ouroboros explore different retrieval or matching mechanisms for candidate generation~\citep{lookahead-fu-2024,pld-saxena-2023,rest-he-2024,token-recycle-luo-2024,liu2025logitspec,hu2025sam,zhao2024ouroboros}.
However, retrieval-only methods often depend on prompt-local overlap, CPU-side structures, suffix automata, or phrase-level matching, which can limit their speedup and make integration with high-throughput tree verification nontrivial.

\paragraph{Parallel and system-level speculative decoding.}
A separate line of work studies how to overlap or pipeline drafting and verification.
Parallel speculative decoding reduces mutual waiting between the draft and target model by adapting draft length or running draft and verification work concurrently~\citep{liu2024parallel,shen2025speculative, shen2026double}.
Lookahead-style frameworks also try to break strict next-token dependency by constructing multiple candidate branches without an external draft model~\citep{lookahead-fu-2024,pia-lookahead-zhao-2024}.
These methods emphasize pipeline utilization and rollback reduction, while high-concurrency systems focus on the interaction between speculation and batched execution.

\paragraph{Speculative decoding for long context.}
Long-context generation shifts the bottleneck toward KV-cache traffic and attention memory bandwidth.
LongSpec improves long-context lossless speculative decoding through efficient drafting and verification~\citep{yang2025longspec}, while SpecPV studies partial verification for long-context self-speculative decoding~\citep{tan2025specpv}.
Other long-context systems explore sparse KV, partial KV, hierarchical speculation, or cache compression to reduce memory pressure.

\paragraph{Block and diffusion drafters.}
Recent work also studies non-autoregressive or block-level drafters that reduce the serial cost of proposal generation.
KVShot provides the first systematic study of long-range decay in hidden-state-based speculative drafters, explores KV-cache reuse to improve long-horizon acceptance, and suggests block-wise paradigms as a promising direction~\citep{liu2026hiddenstatesdriftkv}.
DFlash uses a block diffusion drafter to generate an entire token block in parallel and achieves strong speedups over autoregressive tree drafting on several tasks~\citep{chen2026dflash}.
Nevertheless, block proposals can still suffer from mismatch with target-model autoregressive verification, especially on harder chat or instruction-following data.
Retrieval grafting is a natural complement in this setting: confidence can be used to prune unreliable block tokens, and retrieved continuations can fill the released budget with candidates supported by local history.

\paragraph{Multimodal speculative decoding.}
Speculative decoding for multimodal and vision-language models introduces additional challenges beyond text-only LLMs.
The visual prefix can be long, heterogeneous, and expensive to encode, while the acceptance behavior depends on both visual alignment and text continuation quality.
Recent surveys and systems identify efficient inference for large vision-language models as an emerging bottleneck~\citep{zhang-etal-2026-efficient}.
SpecVLM~\citep{ji-etal-2025-specvlm} is the first to explore training-free speculative decoding for Video-LLMs through vision-aware token pruning on the draft side.
LVSpec~\citep{ji2026foresttreeslooselyspeculative} further extends this line of work by introducing vision-aware loose verification for Video-LLMs.
ParallelVLM extends lossless acceleration to video-LLMs by considering visual-alignment-aware parallel speculative decoding~\citep{kong2026parallelvlm}.
These works suggest that future speculative decoding methods must account for modality-specific proposal quality, visual-token compression, and cross-modal cache reuse.
\end{document}